\documentclass{IEEEtran}
\usepackage{graphicx} 
\usepackage{hyperref}
\usepackage{url}
\usepackage[square,sort&compress,numbers]{natbib}
\usepackage{amsmath}
\usepackage{amssymb}
\usepackage{array}
\usepackage{color}
\usepackage{booktabs}
\usepackage{array, caption, threeparttable}
\usepackage[font=small,labelfont=bf,labelsep=none]{caption}

\usepackage{stfloats}
\usepackage{subfigure}
\title{Boosting Edge Detection with Pixel-wise Feature Selection: The Extractor-Selector Paradigm}
\author{Hao Shu\thanks{Hao Shu: Hao$\_$B$\_$Shu@163.com}\IEEEauthorrefmark{1}
\\
\IEEEauthorblockA{\IEEEauthorrefmark{1} Sun-Yat-Sen University, Shenzhen, China.}
\\
\IEEEauthorblockA{\IEEEauthorrefmark{1} Shenzhen University, Shenzhen, China.}
}
\date{}

\begin{document}
\maketitle

\begin{abstract}

 Deep learning has significantly advanced image edge detection (ED), primarily through improved feature extraction. However, most existing ED models apply uniform feature fusion across all pixels, ignoring critical differences between regions such as edges and textures. To address this limitation, we propose the Extractor-Selector (E-S) paradigm, a novel framework that introduces pixel-wise feature selection for more adaptive and precise fusion. Unlike conventional image-level fusion that applies the same convolutional kernel to all pixels, our approach dynamically selects relevant features at each pixel, enabling more refined edge predictions. The E-S framework can be seamlessly integrated with existing ED models without architectural changes, delivering substantial performance gains. It can also be combined with enhanced feature extractors for further accuracy improvements. Extensive experiments across multiple benchmarks confirm that our method consistently outperforms baseline ED models. For instance, on the BIPED2 dataset, the proposed framework can achieve over 7$\%$ improvements in ODS and OIS, and 22$\%$ improvements in AP, demonstrating its effectiveness and superiority.

\end{abstract}

\section{Introduction}
\label{Introduction}

Image edge detection (ED) is a fundamental task in computer vision, supporting a wide range of high-level applications such as image inpainting\cite{NN2019EdgeConnect}, object detection\cite{ZD2007An}, and image segmentation\cite{MR2011Edge}. Early approaches, including the Sobel operator\cite{K1983On} and the Canny algorithm\cite{C1986A}, utilized gradient-based criteria to develop effective edge detectors. Subsequently, statistical methods with hand-crafted features were proposed\cite{KY2003Statistical, AM2011Contour} to enhance performance. With the maturation of machine learning research, learning-based algorithms have gained significant traction\cite{MF2004Learning, DT2006Supervised, R2008Multi, LZ2013Sketch, GL2014N, DZ2015Fast}. In particular, convolutional neural network (CNN)-based models have rapidly established themselves as state-of-the-art (SOTA) approaches\cite{BS2015Deepedge, LC2017Richer, LK2021Revisiting, SL2021Pixel}, owing to their strong feature extraction capabilities. Notable examples such as Holistically-Nested Edge Detection (HED)\cite{XT2015Holistically}, Bi-Directional Cascade Network (BDCN)\cite{HZ2022BDCN}, and Dense Extreme Inception Network (Dexi)\cite{SS2023Dense} have demonstrated impressive performance by effectively handling complex edge structures. More recently, transformer-based\cite{PH2022Edter, JG2024EdgeNAT} and diffusion-based\cite{ZH2024Generative, YX2024Diffusionedge} architectures have also been explored, though they remain less mature than their CNN-based counterparts. 

A well-performing ED model should exhibit several critical properties. First, it must achieve high precision on established benchmarks, providing numerical validation of its effectiveness. Second, it should produce perceptually accurate results. Third, and importantly, it should perform well without relying on post-processing techniques such as non-maximum suppression (NMS), which are non-differentiable and hinder end-to-end optimization in downstream tasks.

A challenge in ED lies in the inherent ambiguity of defining edges, making it difficult to devise reliable hand-crafted detection criteria. For instance, classical methods like the Canny algorithm\cite{C1986A} rely on gradient features, but textural regions can also exhibit strong gradients despite not corresponding to true edges. This ambiguity demonstrates the importance of comprehensive feature-extracting schemes and necessitates learning-based approaches that extract various cues from annotated datasets. Consequently, CNN-based models\cite{XT2015Holistically, LC2017Richer, HZ2022BDCN, SS2023Dense, LZ2025Cycle} have become dominant in this context due to their superior feature extraction capabilities. 

However, despite their strengths in feature extraction, most CNN-based ED models underexplore the importance of feature fusion. These models typically extract multi-scale features using tens of millions of parameters, yet fuse them using overly simplistic mechanisms—often a single $1\times1$ convolution layer. This uniform fusion does not distinguish between edge and non-edge pixels. Still, in reality, effective feature fusion should be pixel-adaptive: edge pixels benefit from high-resolution features for precise delineation, while texture pixels should emphasize low-level features for suppression purposes. As a result, the indiscriminate fusion currently employed leads to suboptimal predictions. If the feature fusion could be more diverse and context-aware, detection performance would likely improve. 

Motivated by this observation, we propose a novel feature \textbf{Extractor-Selector (E-S)} paradigm to enhance ED through a pixel-wise feature selection mechanism. This paradigm enables adaptive fusion tailored to individual pixels, treating edge and texture regions differently—unlike traditional uniform fusion schemes that fuse all pixels via the same convolutional kernel. The E-S framework consists of two key components: a multi-scale feature extractor that provides diverse feature representations, and a pixel-wise selector that learns to weight these features appropriately. To ensure compatibility with existing ED architectures, the extractor adopts existing CNN-based models, leveraging their proven strengths, while the selector is newly designed as an independent module since no existing model is available. Integration is seamless—existing extractors can be used without modification or with minimal adaptation, yielding significant performance gains either way. Through this pixel-level selection, our approach significantly enhances the accuracy and perceptual quality of edge predictions, as validated by comprehensive experiments. 

Our main contributions can be summarized as follows:

\begin{itemize}
    \item We identify a fundamental limitation in the uniform feature fusion mechanisms used by previous ED models and address it by introducing the Extractor-Selector (E-S) paradigm. 
    \item We propose a pixel-wise feature selector and seamlessly integrate it with existing ED models that serve as feature extractors. The resulting E-S framework achieves state-of-the-art performance.
    \item We further enhance the E-S framework by modifying extractors to support richer feature selection and mitigate information loss, leading to additional improvements in prediction quality. The enhanced version of the E-S framework is called the \textbf{EES} framework.
    \item We conduct extensive experiments on multiple benchmark datasets, demonstrating consistent improvements over models without the E-S framework, highlighting its effectiveness and superiority.
    \item We show that the improvements stem from the E-S paradigm itself, rather than from specific choices of the selector architecture, underscoring the general applicability and robustness of our approach.
\end{itemize}

\section{Previous Works}
\label{Previous Works}

This section reviews related works relevant to the study. 

\subsection{Datasets}

In the early stages of ED research, the task was often conflated with related problems such as contour detection, boundary detection, and segmentation. As a result, early studies typically employed datasets that were not specifically designed for ED but adapted from other tasks. A notable example is the Berkeley Segmentation Dataset BSDS300 and its extended version BSDS500\cite{MF2001A}, which contain 300 and 500 RGB images, respectively, annotated with contours by multiple human annotators. These datasets are primarily used for contour detection today. Similarly, the Multi-cue Boundary Dataset (MDBD)\cite{MK2016A} provides 100 high-resolution images annotated with boundary information across diverse scenes and is commonly used for boundary detection. The NYU Depth Dataset (NYUD) and its extension NYUD2\cite{SH2012Indoor} offer hundreds of thousands of RGB-D indoor scene images, including 1,449 images labeled for segmentation. Other commonly used datasets include PASCAL VOC\cite{EV2010The}, Microsoft COCO\cite{LM2014Microsoft}, SceneParse150\cite{ZZ2017Scene}, and Cityscapes\cite{CO2016The}. While these datasets have significantly contributed to the advancement of ED, their primary focus on other vision tasks limits their specificity and suitability for modern ED research. 

As ED evolved into a distinct research field, specialized datasets were introduced to address its unique requirements. The BIPED dataset, later refined into BIPED2\cite{SR2020Dense}, contains 250 high-resolution images with single-edge annotations. The BRIND dataset\cite{PH2021RINDNet} re-annotates BSDS500 by categorizing edges into four types: reflectance, illumination, normal, and depth. The UDED dataset\cite{SL2023Tiny}, introduced in 2023, consists of 29 carefully selected images with high-quality edge annotations, further enriching the ED benchmark suite, though its limited scale constrains its broader applicability. 

A persistent challenge across ED datasets is their reliance on human annotations for ground truth labeling. These annotations are inherently noisy, often including missing edges, errors, or inconsistencies. Annotation uncertainty also arises when different annotators produce divergent results for the same image—or even when the same annotator produces different annotations at different times. To address these challenges, recent efforts have focused on improving annotation quality using advanced labeling techniques and developing models that are robust to noisy annotations\cite{FG2023Practical, WD2024One, S2025More}. 

\subsection{Models}

The development of ED methodologies can be broadly divided into three phases: heuristic-based methods, statistical approaches, and deep learning-based models.

\textbf{Heuristic-Based Methods:}  
Early ED methods employed simple heuristic criteria, such as intensity gradients, to design efficient edge detectors. Classical techniques such as the Sobel operator\cite{K1983On} and the Canny detector\cite{C1986A} remain widely known due to their simplicity and effectiveness. However, these methods exhibit limited adaptability to complex datasets, as their fixed coefficients struggle to distinguish true edges from textures in diverse scenes. 

\textbf{Statistical and Early Learning-Based Methods:}  
The second phase introduced statistical and early learning-based techniques that improved performance using hand-crafted features. These included Chernoff information\cite{KY2003Statistical}, histograms\cite{AM2011Contour}, textures\cite{MF2004Learning}, and sketch tokens\cite{LZ2013Sketch}. Methods such as structured forests\cite{DZ2015Fast}, nearest-neighbor search\cite{GL2014N}, and logistic regression\cite{R2008Multi} were used to classify pixels or patches as edge or non-edge. These approaches offered improved accuracy and robustness over heuristic methods, but their reliance on manual feature design limited scalability and generalization to more complex datasets. Consequently, they were gradually outpaced by deep learning approaches.

\textbf{Deep Learning-Based Methods:}  
The third and current phase is dominated by deep learning, particularly convolutional neural networks (CNNs). CNN-based ED models have revolutionized the field by enabling hierarchical feature extraction and end-to-end learning. Holistically-Nested Edge Detection (HED)\cite{XT2015Holistically} introduced multi-scale learning by fusing features from different scales to enhance edge prediction. Richer Convolutional Features (RCF)\cite{LC2017Richer} extended this by aggregating features from multiple stages within the same layer. Bi-Directional Cascade Networks (BDCN)\cite{HZ2022BDCN} and Pixel Difference Networks (PiDiNet)\cite{SL2021Pixel} introduced novel architectures to improve edge precision. RINDNet\cite{PH2021RINDNet} addressed multiple edge types through multi-stage fusion, while DexiNet\cite{SS2023Dense} leveraged dense skip connections for better accuracy. More recently, transformer-based\cite{PH2022Edter, JG2024EdgeNAT} and diffusion-based\cite{ZH2024Generative, YX2024Diffusionedge} models have emerged, offering promising alternative paradigms, though they are still less mature than CNN-based methods. 

\subsection{Loss Functions}

Most ED models rely on weighted binary cross-entropy (WBCE) loss, which promotes high precision but struggles to produce crisp, thin edges without post-processing techniques such as non-maximum suppression (NMS). To overcome this limitation, alternative loss functions such as Dice loss\cite{DS2018Learning} and tracing loss\cite{HX2022Unmixing} have been proposed to produce thinner edges. However, these often involve trade-offs in precision or require careful hyperparameter tuning across datasets and models. Another key issue is the class imbalance between edge and non-edge pixels, which can degrade model performance. In response, methods such as AP-loss\cite{CL2019Towards} and Rank-loss\cite{ CK2024RankED} have been introduced. However, the WBCE loss is still the most popular one currently. 

\subsection{Encoder-Decoder Framework in Transformers}

Originally developed for natural language processing\cite{VS2017Attention}, the encoder-decoder framework has proven effective for image processing tasks as well\cite{D2020An}. This structure captures long-range dependencies without necessitating excessively deep architectures, in contrast to traditional CNNs. Transformer-based models leveraging this framework have shown success in tasks such as object detection\cite{CM2020End}, semantic segmentation\cite{ZL2021Rethinking}, and image super-resolution\cite{ZZ2025HiT}. Recently, transformers have also been applied to contour detection\cite{PH2022Edter}, expanding their utility within the ED domain. 

\subsection{Evaluation Benchmarks}

The most widely adopted evaluation protocol for ED follows the methodology proposed in \cite{MF2004Learning}, using three standard metrics: 

\begin{itemize}
    \item \textbf{Optimal Dataset Score (ODS):} Measures the best F-score across the entire dataset.
    \item \textbf{Optimal Image Score (OIS):} Computes the best F-score per image, then averages the scores.
    \item \textbf{Average Precision (AP):} Evaluates the mean precision-recall performance.
\end{itemize}

These metrics are computed under a predefined error tolerance distance, which allows slight misalignments between predicted and ground-truth edges. However, the choice of error tolerance often varies across datasets and studies due to reliance on default parameters. Limited attention has been given to standardizing this parameter until a recent study analyzed its effect and advocated for a 1-pixel tolerance to provide the most rigorous evaluation benchmark\cite{S2025More}. 

\section{Methodology}

This section presents our proposed approach. We will begin by analyzing the limitations in existing ED models. Then, we will introduce the standard Extractor-Selector (E-S) paradigm and a selector model for achieving it, followed by an enhanced version—the Enhanced E-S (EES) framework—designed to improve feature quality and reduce information loss to further improve the final edge predictions. Finally, we will describe the training strategies used for the proposed schemes.

\subsection{The challenge in Existing Models}

Conventional ED models can be divided into two stages: feature extraction and feature fusion. While significant effort is dedicated to building powerful extractors—often comprising tens of millions of parameters—the feature fusion step is often overly simplistic, commonly implemented as a single 1$\times$1 convolution layer. This imbalance limits the models’ ability to leverage the rich multi-scale features extracted in the first stage.

Essentially, feature fusion in ED is inherently a feature selection problem. Effective edge detection requires high-resolution features for localizing edge pixels precisely and low-resolution features for suppressing texture noise in non-edge regions. However, a 1$\times$1 convolution layer applies the same weights across all spatial locations, failing to distinguish between edge and non-edge regions. Consequently:

\begin{itemize}
    \item If high-resolution features are prioritized by the fusion layer, texture regions may be misclassified as edges (false positives).
    \item If low-resolution features are prioritized by the fusion layer, edges may become blurry or inaccurate (false negatives).
\end{itemize}

Thus, even with globally optimized weights, this uniform treatment limits the model’s ability to perform well across all pixel types.

As demonstrated in \cite{S2025More}, enhancing the fusion process yields performance gains. However, simply stacking additional convolutional layers as in the paper only marginally alleviates the issue because it fails to resolve the underlying problems stated above. The limited capacity of shallow convolutional fusion modules is insufficient to fully exploit the rich feature sets produced by modern extractors, nor to provide differentiated feature selections between edge and texture pixels.

This motivates the introduction of a dedicated feature selection module capable of assigning pixel-wise adaptive weights to multi-scale features—allowing for more discriminative and precise feature selection for edge predictions.

\subsection{The Standard E-S Architecture}

To address the above challenge, we propose the E-S framework. It preserves traditional ED models as feature extractors but introduces a dedicated feature selector that assigns pixel-wise weights to multi-scale features for selecting the extracted features, thereby enhancing both accuracy and perceptual quality.

The E-S framework consists of two components: a feature extractor that captures multi-scale features from input images, and a feature selector that assigns pixel-wise weights to extracted features to generate refined edge maps. An overview of the framework is shown in Figure \ref{Extractor-Selector-Paradigm}.

\begin{figure*}[htbp]
	\centering
	\includegraphics[width=7in]{Extractor_Selector_Paradigm.png}
	\caption{\quad\textbf{The framework of the E-S paradigm.} It consists of two components: a feature extractor which produces features and coarse edge maps, and a feature selector which selects the features for refinements. The combination of the two components ultimately outputs the final edge predictions.}
	\label{Extractor-Selector-Paradigm}
\end{figure*}

The extractor can adopt any existing high-performance ED model—such as HED\cite{XT2015Holistically}, BDCN\cite{HZ2022BDCN}, Dexi\cite{SS2023Dense}, or even their combination, to leverage their superior feature extraction capabilities. However, because no off-the-shelf selector model exists, a custom selector architecture is required, which will be introduced in the next subsection.

Generally, the E-S framework can be formulated as:

\begin{equation}
\begin{aligned}
    Edge(I)=FSM(P_{2}(E_{\theta}(I)),S_{\eta}(I))
\end{aligned}
\end{equation}

\begin{equation}
\begin{aligned}
    FSM:\ &\mathbb{R}^{C\times H\times W}\times \mathbb{R}^{C\times H\times W}\ \longrightarrow \qquad \mathbb{R}^{1\times H\times W}\\
    &\qquad\ \ (\ A\ ,\  B\ )\qquad\ \ \ \longmapsto\quad\sum_{dim:\ 1}(A\odot B)
\end{aligned}
\end{equation}

\begin{equation}
\begin{aligned}
    E_{\theta}: \mathbb{R}^{3\times H\times W}\quad\ &\longrightarrow \qquad\quad \mathbb{O}\ \ \times\ \  \mathbb{R}^{C\times H\times W}\\
    I\quad\qquad&\longmapsto \ \ (P_{1}(E_{\theta}(I)),P_{2}(E_{\theta}(I)))
\end{aligned}
\end{equation}

\begin{equation}
\begin{aligned}
    S_{\eta}: \mathbb{R}^{3\times H\times W}\qquad&\longrightarrow \qquad \mathbb{R}^{C\times H\times W}\\
    I\qquad\qquad&\longmapsto \qquad\quad S_{\eta}(I)
\end{aligned}
\end{equation}

\noindent where $P_{1}$ and $P_{2}$ denote the projections onto the first and second components of a direct product, respectively. $I$ represents the input image. The symbol $\odot$ denotes point-wise multiplication, and $\sum_{dim: 1}$ denotes summation over the first dimension (i.e. the channel dimension). The functions $E_{\theta}$ and $S_{\eta}$ refer to the networks of the feature extractor and feature selector, respectively, parameterized by $\theta$ and $\eta$, both outputting $C$ channels from an input 3-channel (RGB) image. $\mathbb{O}$ represents the space of edge maps produced by the extractor, while the second output component of $E_{\theta}$ corresponds to its outputted feature. Here, $H$ and $W$ are the height and width of the input image.

In the standard implementation, the selector assigns pixel-wise weights directly to the final multi-scale feature maps outputted by the extractor to produce the final edge prediction. Thus, $P_{1}(E_{\theta}(I))$ corresponds to the extractor’s original edge map outputs, typically as a list, while $P_{2}(E_{\theta}(I))$ is their concatenation. Moreover, a slight modification to the extractor can further enhance performance, where $P_{1}(E_{\theta}(I))$ and $P_{2}(E_{\theta}(I))$ are significantly different. This improved design will be discussed in detail two sections later.

\subsection{Design of the Feature Selector}

\begin{figure*}[htbp]
	\begin{center}
		\centerline{\includegraphics[width=1.8\columnwidth]{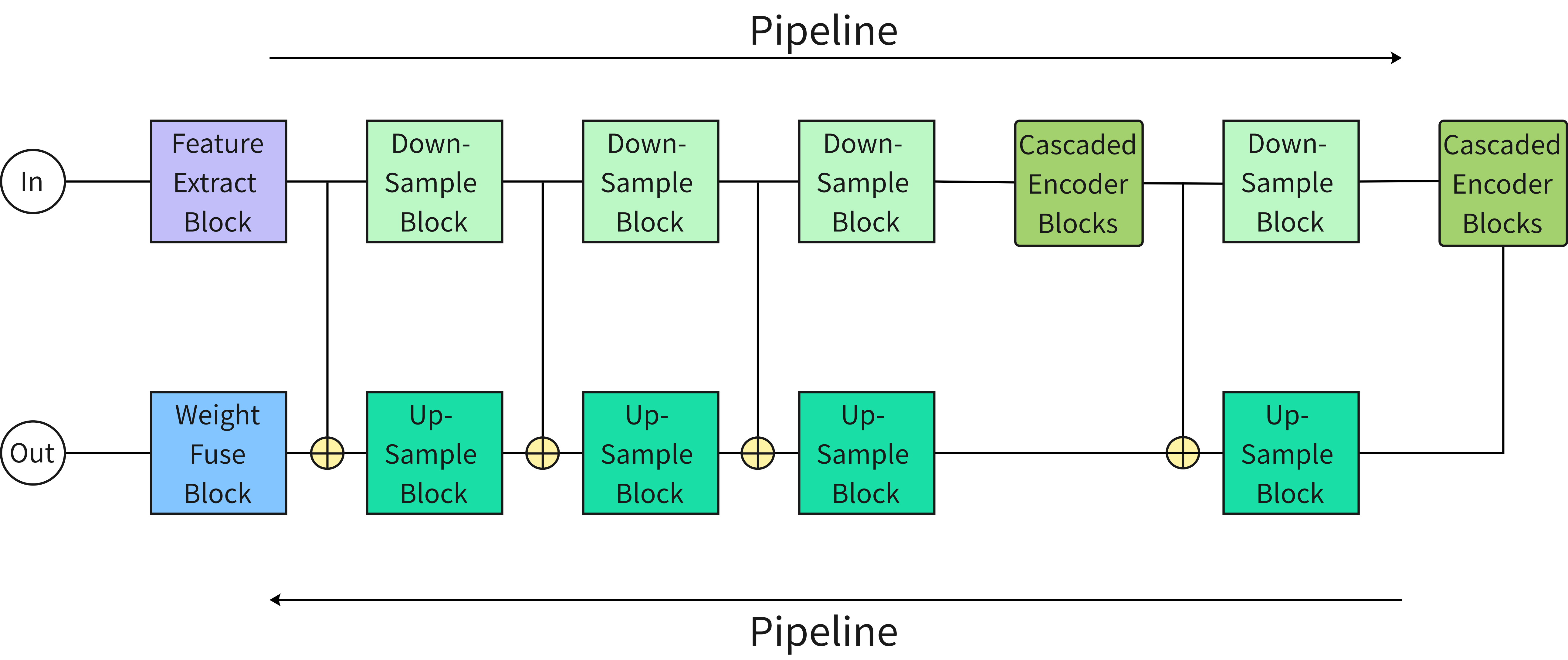}}
		\caption{\quad\textbf{An overview of the feature selector architecture:} It is structured like a U-Net. The input images pass through a feature extraction block, followed by four $\frac{1}{2}$ down-sampling blocks to capture multi-scale features. At the $\frac{1}{8}$ and $\frac{1}{16}$ scales, cascaded transformer encoder blocks are applied. The features from the $\frac{1}{16}$ scale are progressively up-sampled, and each up-sample block incorporates a residual connection with a learnable coefficient to balance the up-sampled features and the features from the residual connections. Finally, pixel-wise weights are produced by fusing the ordinary-scale features. Detailed block designs are presented in supplementary materials.}
		\label{Selector}
	\end{center}
\end{figure*}

The selector should be selected with some key properties, including high expressive capacity to enable diverse and precise feature selection and adequate degrees of freedom to generate pixel-wise adaptive weights. Since these properties align with established deep learning paradigms, we employ a deep U-Net-inspired architecture for adequate degrees of freedom and combine CNNs with transformer encoders for high expressive capacity. CNNs excel at capturing short-range dependencies, which are crucial for local edge refinement, while transformers effectively model long-range dependencies using a relatively shallow structure. Combining these architectures ensures both local precision and global consistency and therefore enriches the expressive capacity of the architecture. 

An overview of the selector structure is shown in Figure \ref{Selector}. It follows a U-Net-like CNN backbone, and at the $\frac{1}{8}$ and $\frac{1}{16}$ scales, transformer-based encoder-decoder structures are integrated. The final fusion stage employs a three-layer architecture, similar to SDPED\cite{S2025More}, which has been proven to be effective. Additionally, trainable coefficients are added to the residual connections to optimize the balance between the output features and those linked through the residual connections.

\subsection{The EES Architecture}

While the standard E-S framework enables pixel-wise selection, it only operates on the final fused outputs from the extractor. These features, having passed through several compression layers, often lose valuable detail. Inspired by RCF\cite{LC2017Richer}, which enriches feature utilization in HED\cite{XT2015Holistically}, we propose an Enhanced framework, namely the EES framework, that exposes the selector to richer, less compressed features from earlier stages of the extractor.

\begin{figure}[htbp]
	\centering
	\includegraphics[width=0.45\textwidth]{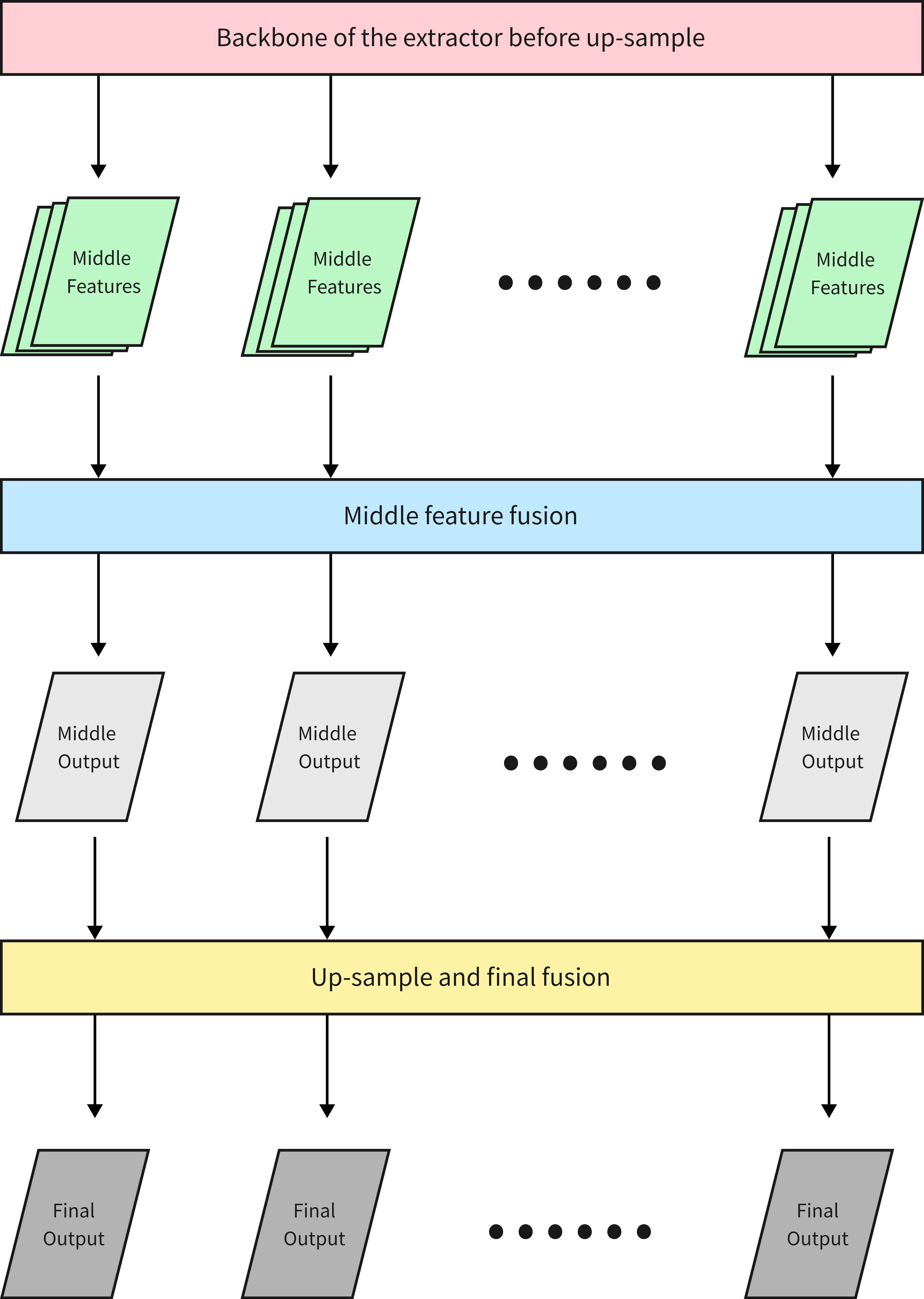}
	\caption{\quad \textbf{The standard extractor architecture:} The backbone network extracts multi-scale features through progressive down-sampling. These features are then fused into one or two edge maps at each scale and then up-sampled to the original image resolution. The final edge prediction is obtained by fusing these up-sampled maps. However, because the features are compressed prior to up-sampling, the resulting high-resolution representations often suffer from substantial information loss, limiting the usage by the selector.}
	\label{Old-Extractor}
\end{figure}

\begin{figure}[htbp]
	\centering
	\includegraphics[width=0.5\textwidth]{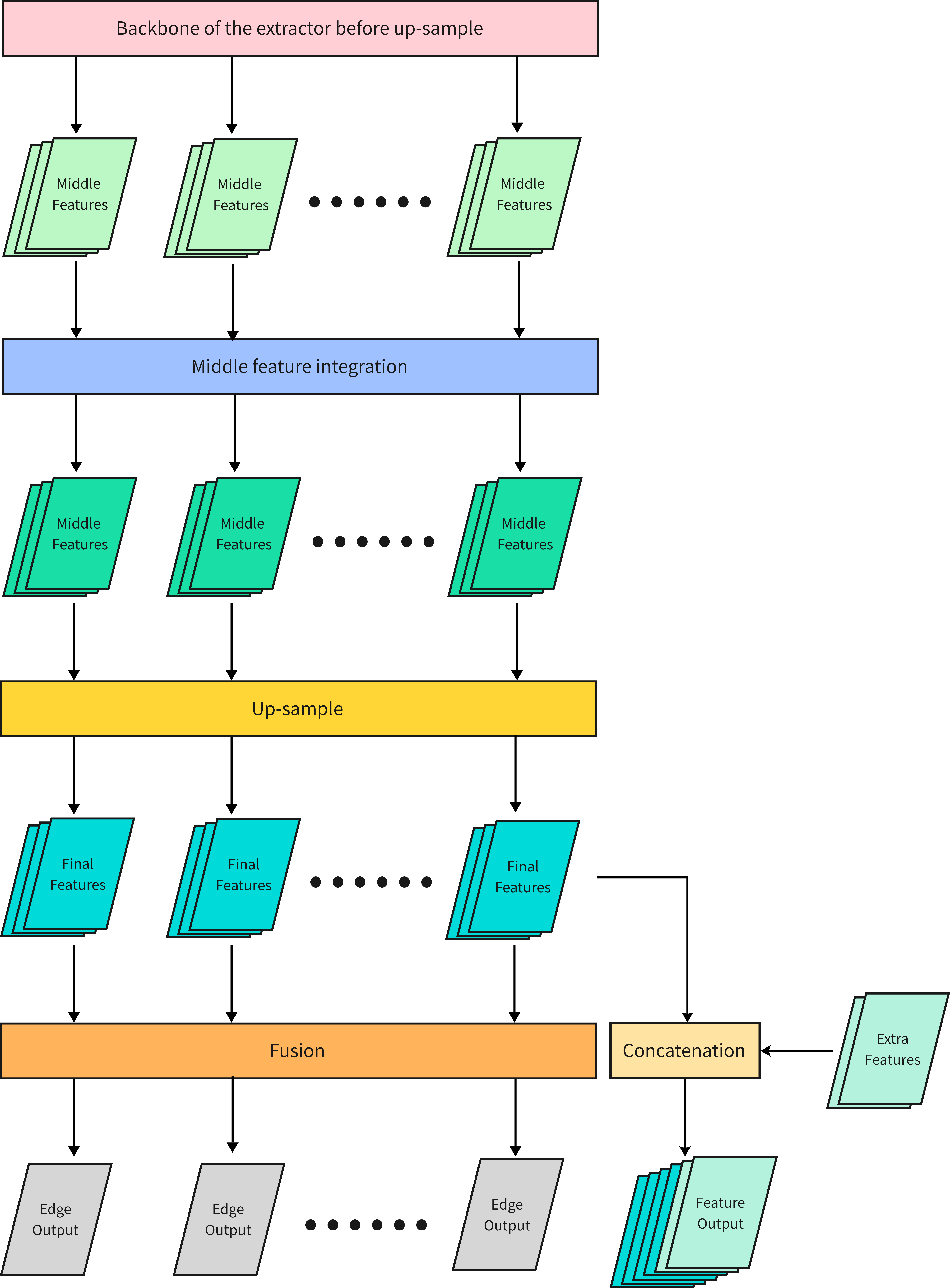}
	\caption{\quad\textbf{The modified feature extractor architecture.} Unlike the standard extractor architecture, which compresses multi-scale features before up-sampling, the modified version preserves richer information by avoiding early fusion. It uses the same backbone to extract multi-scale features, which are first unified to a suitable number of channels and then up-sampled without compression. Two additional fixed feature maps (one filled with zeros and one with ones) are appended to assist the selector in identifying highly confident edge and non-edge regions. Although a coarse fusion is applied during pre-training of the extractor, it is these enhanced, unfused features that are passed to the selector for final prediction, providing more refined and less lossy representations compared to the standard approach.}
	\label{New-Extractor}
\end{figure}

To this end, we modify the extractor’s fusion and upsampling modules, without altering its backbone, as shown in Figures \ref{Old-Extractor} and \ref{New-Extractor}. 

Specifically, previous ED models (Figure \ref{Old-Extractor}), which serve as extractors in the E-S framework, typically consist of a backbone network with down-sampling blocks that progressively reduce the spatial resolution. Multi-scale features are then obtained, fused to one or two edge maps in each scale, and passed through up-sampling blocks to generate final features, at the target resolution, which are further fused to produce the final edge map. Although this structure has been effective in many cases, the final features often suffer from significant information loss and lack of diversity due to the compressed fusion schemes before up-sampling.

In contrast, the modified architecture (Figure \ref{New-Extractor}) integrates multi-scale features into a suitable number of channels, which are then up-sampled without fusion. The up-sampled features are used as the output, offering richer information with reduced loss for selection. Although the final outputted features are still fused into coarse edge maps for pre-training the extractor, it is the unfused features that are used by the selector for the final predictions. Additionally, we introduce two fixed feature maps (one of zeros and one of ones) to assist the selector in making more confident decisions, particularly in cases where a pixel is strongly classified as belonging to either the edge or non-edge region.

\subsection{Training Strategy}

 A three-stage training strategy is adopted:

\begin{itemize}
    \item Stage 1 (Feature Extractor Pretraining): The feature extractor is trained independently via its edge outputs $P_{1}(E_{\theta}(I))$, using its default loss function $L_{E}$. Namely,
    \begin{equation}
    \theta_{E}=\underset{\theta^{*}}{\arg min}
\ L_{E}(P_{1}(E_{\theta}(I)),Label(I))
    \end{equation}
    \noindent where $Label(I)$ denotes the ground-truth of the input image $I$.
    \item Stage 2 (Feature Selector Training): The feature extractor’s parameters are frozen while the feature selector is trained separately using loss function $L_{S}$, typically the WBCE loss. Namely,
    \begin{equation}
    \eta_{S}=\underset{\eta^{*}}{\arg min}
\ L_{S}(FSM(P_{2}(E_{\theta_{E}}(I),S_{\eta^{*}}(I))),Label(I))
    \end{equation}
    \item Stage 3 (Union Tuning): Both the feature extractor and selector are trained jointly with loss function $L_{U}$, typically WBCE. Namely,
    \begin{equation}
    \theta, \eta=\underset{\theta^{*},\eta^{*}}{\arg min}
\ L_{U}(FSM(P_{2}(E_{\theta^{*}}(I),S_{\eta^{*}}(I))),Label(I))
    \end{equation}
\noindent where the initial $\theta^{*}$ and $\eta^{*}$ are set to $\theta_{E}$ and $\eta_{S}$, respectively.
\end{itemize}

\vspace{0.5cm}
The WBCE loss is defined as:
\begin{small}
\begin{equation}
	 L_{WBCE}(\hat{Y},Y)=\alpha\sum_{y_{i}\in Y^{+}}log(\hat{y_{i}})-\lambda(1-\alpha)\sum_{y_{i}\in Y^{-}}log(1-\hat{y_{i}})
\end{equation}
\end{small}

\noindent where:
\begin{itemize}
	\item $\hat{y_{i}}$ is the pixel in the prediction $\hat{Y}$ corresponding to $y_{i}$.
	\item $Y^{+}$ represents the set of edge pixels in the ground-truth $Y$ (positive samples).
	\item $Y^{-}$ denotes the set of non-edge pixels in the ground-truth (negative samples).
	\item $\alpha = \frac{|Y^{-}|}{|Y|}$ is the weight balancing the positive and negative samples.
	\item $\lambda = 1.1$ as suggested in previous works.
\end{itemize}

\section{Experiment}

This section presents our experimental setup, results, and ablation studies to validate the effectiveness of the proposed frameworks.

\subsection{Experimental Setup and Benchmarks}

The experimental setup follows the methodology outlined in \cite{MF2004Learning} and refined in \cite{S2025More}. Below, we provide detailed descriptions of the datasets, augmentation techniques, training procedures, and evaluation criteria.

\subsubsection{Datasets and Data Augmentation}

The main experiments are conducted on three ED datasets: BRIND\cite{PH2021RINDNet}, UDED\cite{SL2023Tiny}, and BIPED2\cite{SR2020Dense}. Additional experiments are available in the supplementary material. The preprocessing of the dataset is stated as follows.

\textbf{Label employments:}
\begin{itemize}
    \item For BRIND, all annotated edge maps are merged into a single binary ground truth map per image. A pixel is marked as an edge if any annotator labeled it as such.
    \item UDED and BIPED2 provide single, unique ground truth edge annotations per image, which are used directly.
\end{itemize}

\textbf{Dataset splits:}
\begin{itemize}
    \item BRIND: 500 images, split into 400 training and 100 test samples.
    \item BIPED2: 250 images, split into 200 training and 50 test samples.
    \item UDED: 27 usable images (after excluding two with height or width lower than 320 pixels), split into 20 training and 7 test samples.
\end{itemize}

\textbf{Data augmentation (follows \cite{S2025More}):}
\begin{itemize}
    \item Each image is recursively split in half until its height and width are both lower than 640 pixels.
    \item Augmentation includes rotations (0°, 90°, 180°, and 270°) and horizontal flips.
    \item Noiseless data are added.
\end{itemize}

\subsubsection{Training Notes}

During training, 320$\times$320 pixel crops are sampled randomly and refreshed every 5 epochs. All models are optimized using Adam with a learning rate of $10^{-4}$, weight decay of $10^{-8}$, and batch size of 8. The training procedure is stated as follows.

\textbf{Training process:}
\begin{itemize}
    \item Stage 1 (Feature Extractor Pretraining): The extractor is trained independently, with 50 epochs for the BRIND and BIPED2, and 200 epochs for the UDED, respectively.
    \item Stage 2 (Feature Selector Training): The selector is trained, freezing the extractor parameters, with an additional 50 epochs for BRIND and BIPED2, and 200 epochs for UDED, respectively.
    \item Stage 3 (Union Stage): Both the extractor and the selector are trained jointly, with another 50 epochs for BRIND and BIPED2, and 200 epochs for UDED, respectively.
\end{itemize}

\subsubsection{Evaluation Metrics}

Performance is evaluated using the framework from \cite{MF2004Learning}, with a strict 1-pixel error tolerance as suggested in \cite{S2025More}. This is in contrast to most prior works which used error tolerances of 4 to 11 pixels, and the reason for using 1-pixel error tolerance has been discussed in \cite{S2025More}\footnote{The 1-pixel error tolerance corresponds to setting the error toleration distance to 0.001 for BIPED2 and 0.003 for BRIND and UDED in the MATLAB evaluation code from \cite{MF2004Learning}.}. All results are reported without post-processing, such as non-maximum suppression, whose reason has been illustrated in the introduction. Results with traditional relaxed error tolerances with post-processing are provided in the supplementary material. To assess the effectiveness of the proposed frameworks: 

\textbf{We evaluate:}

\begin{itemize}
    \item The original extractor models (previous models), including HED\cite{XT2015Holistically}, BDCN\cite{HZ2022BDCN}, and Dexi\cite{SS2023Dense}). They serve as baseline models.
    \item The corresponding E-S versions (of each baseline model) without joint training (denoted -ES).
    \item The corresponding E-S versions with joint training (denoted -ES-U).
    \item The corresponding EES versions without joint training (denoted -EES).
    \item The corresponding EES versions with joint training (denoted -EES-U).
\end{itemize}

\textbf{For fairness:}
\begin{itemize}
    \item All models are trained or re-trained under identical settings given above.
    \item All models are trained with WBCE loss, which was also used as the default loss function of the baseline models.
    \item All results undergo the same prediction Procedure: Test images are processed in overlapping 320$\times$320 patches with 16-pixel overlap.\footnote{If the last patch in a row or column is smaller than 304 pixels, additional overlap is applied to the last two patches in that row or column to ensure the last patch is of 320$\times$320. Details can be found in 
 the implementation codes.} Overlapping regions are averaged to ensure spatial consistency in the final output.
\end{itemize}

\subsection{Experiment Results}

Results for the BIPED2, BRIND, and UDED datasets (under 1-pixel error tolerance and without NMS) are summarized in Tables \ref{Raw-BIPED2}, \ref{Raw-UDED}, and \ref{Raw-BRIND}, respectively. They include the performance of various feature extractors and their combinations with feature selectors, both the standard and the enhanced versions. The improvements relative to the baseline models are explicitly indicated.

\textbf{Quantitative Analysis}

In all the benchmark datasets and all the assessing criteria, both the EES and the standard E-S architectures achieve significant improvements. The average improvements over the three extractors, compared to the baseline models, in ODS, OIS, and AP, respectively, are summarized as follows.

\textbf{For BIPED2:}
\begin{itemize}
    \item Standard E-S framework without union training improves: 3.72$\%$, 3.53$\%$, and 8.00$\%$, respectively.
    \item Standard E-S framework with union training improves: 6.02$\%$, 6.09$\%$, and 20.00$\%$, respectively.
    \item EES framework without union training improves: 7.28$\%$, 7.05$\%$, and 22.00$\%$, respectively.
    \item EES framework with union training improves: 5.83$\%$, 5.93$\%$, and 18.00$\%$, respectively.
\end{itemize}

\textbf{For UDED:}
\begin{itemize}
    \item Standard E-S framework without union training improves: 1.95$\%$, 1.60$\%$, and 4.07$\%$, respectively.
    \item Standard E-S framework with union training improves: 4.60$\%$, 3.07$\%$, and 14.63$\%$, respectively.
    \item EES framework without union training improves: 5.43$\%$, 4.01$\%$, and 20.19$\%$, respectively.
    \item EES framework with union training improves: 6.55$\%$, 4.94$\%$, and 13.52$\%$, respectively.
\end{itemize}

\textbf{For BRIND:}
\begin{itemize}
    \item Standard E-S framework without union training improves: 1.83$\%$, 1.20$\%$, and 5.51$\%$, respectively.
    \item Standard E-S framework with union training improves: 1.83$\%$, 1.35$\%$, and 17.40$\%$, respectively.
    \item EES framework without union training improves: 3.50$\%$, 3.15$\%$, and 10.35$\%$, respectively.
    \item EES framework with union training improves: 1.83$\%$, 1.20$\%$, and 17.40$\%$, respectively.
\end{itemize}

These results clearly demonstrate that both the standard E-S and the enhanced EES frameworks significantly improve edge detection performance across all datasets, and EES mostly obtains more gains, which should be attributed to its richer feature representations.

\begin{table}[htbp]\scriptsize
\renewcommand\arraystretch{1.5}
\centering
\caption{\qquad \textbf{Results on BIPED2 with 1-pixel error tolerance without NMS:} For each extractor, the first row presents the results of the baseline model, the rows \textit{-ES}, \textit{-ES-U}, \textit{-EES}, and \textit{-EES-U} exhibits results with our frameworks, following the notations in the main-text. The best score among the configurations for each extractor is highlighted in \textbf{bold}. Improvements related to the baseline models are indicated in parentheses. The last five rows \textit{(A)} report average improvements across the three baseline models. Improvements are also indicated in parentheses with the best scores in \textbf{bold}.}
\label{Raw-BIPED2}
\begin{tabular}{|p{16mm}<{\centering}|p{65.75mm}<{\centering}|}
\hline
 &  BIPED2 with 1-pixel error tolerance without NMS
\end{tabular}
\begin{tabular}{|p{16mm}<{\centering}|p{19mm}<{\centering}|p{19mm}<{\centering}|p{19mm}<{\centering}|}
\hline
    & ODS   & OIS   & AP \\
\hline
HED & 0.592 & 0.602 & 0.347 \\
\hline
HED-ES & 0.649 (+9.63$\%$) & 0.654 (+8.64$\%$) & 0.433 (+24.78$\%$) \\
\hline
HED-EES & \textbf{0.666 (+12.50$\%$)} & \textbf{0.671 (+11.46$\%$)} & \textbf{0.493 (+42.07$\%$)} \\
\hline
HED-ES-U & 0.655 (+10.64$\%$) & 0.663 (+10.13$\%$) & 0.491 (+41.50$\%$) \\
\hline
HED-EES-U & 0.659 (+11.32$\%$) & 0.665 (+10.45$\%$) & 0.490 (+41.21$\%$) \\
\hline
BDCN & 0.629 & 0.635 & 0.421 \\
\hline
BDCN-ES & 0.633 (+0.64$\%$) & 0.639 (+0.63$\%$) & 0.423 (+0.46$\%$) \\
\hline
BDCN-EES & 0.652 (+3.66$\%$) & 0.655 (+3.15$\%$) & 0.473 (+12.35$\%$) \\
\hline
BDCN-ES-U & \textbf{0.661 (+5.09$\%$)} & \textbf{0.666 (+4.88$\%$)} & \textbf{0.484 (+14.96$\%$)} \\
\hline
BDCN-EES-U & 0.652 (+3.66$\%$) & 0.660 (+3.94$\%$) & 0.462 (+9.74$\%$) \\
\hline
Dexi & 0.632 & 0.636 & 0.432 \\
\hline
Dexi-ES & 0.640 (+1.27$\%$) & 0.644 (+1.26$\%$) & 0.441 (+2.08$\%$) \\
\hline
Dexi-EES & \textbf{0.672 (+6.33$\%$)} & \textbf{0.677 (+6.45$\%$)} & \textbf{0.499 (+15.51$\%$)} \\
\hline
Dexi-ES-U & 0.650 (+2.89$\%$) & 0.656 (+3.14$\%$) & 0.465 (+7.64$\%$) \\
\hline
Dexi-EES-U & 0.651 (+3.01$\%$) & 0.657 (+2.99$\%$)& 0.463 (+7.18$\%$) \\
\hline
A & 0.618 & 0.624 & 0.400 \\
\hline
A-ES & 0.641 (+3.72$\%$) & 0.646 (+3.53$\%$) & 0.432 
 (+8.00$\%$) \\
\hline
A-EES & \textbf{0.663 (+7.28$\%$)} & \textbf{0.668 (+7.05$\%$)} & \textbf{0.488 (+22.00$\%$)} \\
\hline
A-ES-U & 0.655 (+6.02$\%$) & 0.662 (+6.09$\%$) & 0.480 (+20.00$\%$) \\
\hline
A-EES-U & 0.654 (+5.83$\%$) & 0.661 (+5.93$\%$) & 0.472 (+18.00$\%$) \\
\hline
\end{tabular}
\end{table}

\begin{table}[htbp]\scriptsize
\renewcommand\arraystretch{1.5}
\centering
\caption{\qquad \textbf{Results on UDED with 1-pixel error tolerance (for the lowest-resolution images) without NMS:} The table is structured as Table \ref{Raw-BIPED2}, with results for the UDED dataset.}
\label{Raw-UDED}
\begin{tabular}{|p{16mm}<{\centering}|p{65.75mm}<{\centering}|}
\hline
 &  UDED without NMS
\end{tabular}
\begin{tabular}{|p{16mm}<{\centering}|p{19mm}<{\centering}|p{19mm}<{\centering}|p{19mm}<{\centering}|}
\hline
    & ODS   & OIS   & AP \\
\hline
HED & 0.716 & 0.751 & 0.489 \\
\hline
HED-ES & 0.733 (+2.37$\%$) & 0.764 (+1.73$\%$) & 0.521 (+6.54$\%$) \\
\hline
HED-EES & 0.738 (+3.07$\%$) & 0.765 (+1.86$\%$) & \textbf{0.635 (+29.86$\%$)} \\
\hline
HED-ES-U & 0.723 (+0.98$\%$) & 0.751 (+0.00$\%$) & 0.602 (+23.11$\%$) \\
\hline
HED-EES-U & \textbf{0.754 (+5.31$\%$)} & \textbf{0.782 (+4.13$\%$)} & 0.616 (+25.97$\%$) \\
\hline
BDCN & 0.706 & 0.744 & 0.530 \\
\hline
BDCN-ES & 0.727 (+2.97$\%$) & 0.764 (+2.69$\%$) & 0.569 (+7.36$\%$) \\
\hline
BDCN-EES & 0.750 (+6.23$\%$) & 0.777 (+4.44$\%$) & \textbf{0.628 (+18.49$\%$)} \\
\hline
BDCN-ES-U & 0.758 (+7.37$\%$) & \textbf{0.778 (+4.57$\%$)} & 0.625 (+17.92$\%$) \\
\hline
BDCN-EES-U & \textbf{0.762 (+7.93$\%$)} & \textbf{0.778 (+4.57$\%$)} & 0.623 (+17.55$\%$) \\
\hline
Dexi & 0.733 & 0.752 & 0.600 \\
\hline
Dexi-ES & 0.737 (+0.55$\%$) & 0.759 (+0.93$\%$) & 0.598 (-0.33$\%$) \\
\hline
Dexi-EES & \textbf{0.784 (+6.96$\%$)} & 0.796 (+5.85$\%$) & \textbf{0.683 (+13.83$\%$)} \\
\hline
Dexi-ES-U & 0.771 (+5.18$\%$) & 0.786 (+4.52$\%$) & 0.629 (+4.83$\%$) \\
\hline
Dexi-EES-U & 0.780 (+6.41$\%$) & \textbf{0.797 (+5.98$\%$)} & 0.601 (+0.17$\%$) \\
\hline
A & 0.718 & 0.749 & 0.540 \\
\hline
A-ES & 0.732 (+1.95$\%$) & 0.761 (+1.60$\%$) & 0.562 (+4.07$\%$) \\
\hline
A-EES & 0.757 (+5.43$\%$) & 0.779 (+4.01$\%$) & \textbf{0.649 (+20.19$\%$)} \\
\hline
A-ES-U & 0.751 (+4.60$\%$) & 0.772 (+3.07$\%$) & 0.619 (+14.63$\%$) \\
\hline
A-EES-U & \textbf{0.765 (+6.55$\%$)} & \textbf{0.786 (+4.94$\%$)} & 0.613 (+13.52$\%$) \\
\hline
\end{tabular}
\end{table}

    \begin{figure*}[htbp]
        \centering
        \includegraphics[width=7in]{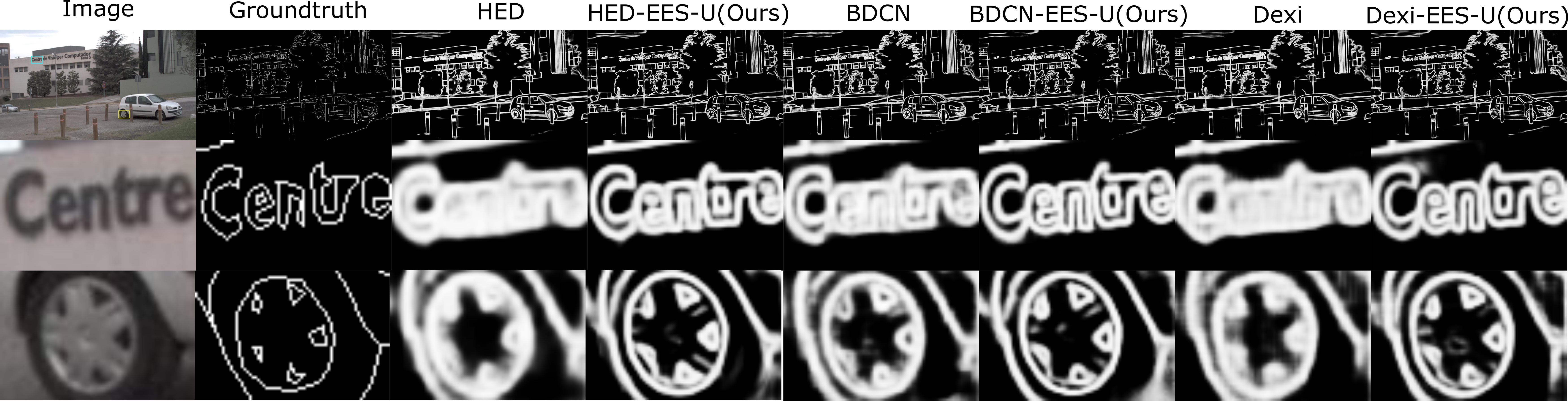}
        \caption{ \quad\textbf{Visual comparisons of the models:} Columns 1 presents the ordinary image and its cropped regions marked by blue (the up-left words) and yellow boxes (the down-right tire), respectively. Column 2 displays the corresponding ground-truths. Columns 3, 5, and 7 display the predictions of the baseline models, while columns 4, 6, and 8 exhibit the predictions of the corresponding EES frameworks. Obviously, the EES frameworks obtain better perceptual results.}
        \label{Compare}
    \end{figure*}

\begin{table}[htbp]\scriptsize
\renewcommand\arraystretch{1.5}
\centering
\caption{\textbf{Results on BRIND with 1-pixel error tolerance without NMS:} The table is structured as in Table \ref{Raw-BIPED2}, with results for the BRIND dataset.}
\label{Raw-BRIND}
\begin{tabular}{|p{16mm}<{\centering}|p{65.75mm}<{\centering}|}
\hline
 &  BRIND without NMS
\end{tabular}
\begin{tabular}{|p{16mm}<{\centering}|p{19mm}<{\centering}|p{19mm}<{\centering}|p{19mm}<{\centering}|}
\hline
    & ODS   & OIS   & AP \\
\hline
HED & 0.645 & 0.656 & 0.417 \\
\hline
HED-ES & 0.668 (+3.57$\%$) & 0.677 (+3.20$\%$) & 0.466 (+11.75$\%$) \\
\hline
HED-EES & \textbf{0.679 (+5.27$\%$)} & \textbf{0.687 (+4.73$\%$)} & 0.492 (+17.98$\%$) \\
\hline
HED-ES-U & 0.668 (+3.57$\%$) & 0.673 (+2.59$\%$) & 0.522 (+25.18$\%$) \\
\hline
HED-EES-U & 0.669 (+3.72$\%$) & 0.676 (+3.05$\%$) & \textbf{0.543 (+30.22$\%$)} \\
\hline
BDCN & 0.659 & 0.672 & 0.467 \\
\hline
BDCN-ES & 0.668 (+1.37$\%$) & 0.677 (+0.74$\%$) & 0.486 (+4.07$\%$) \\
\hline
BDCN-EES & \textbf{0.678 (+2.88$\%$)} & \textbf{0.686 (+2.08$\%$)} & 0.521 (+10.92$\%$) \\
\hline
BDCN-ES-U & 0.672 (+1.97$\%$) & \textbf{0.681 (+1.34$\%$)} & 0.516 (+10.49$\%$) \\
\hline
BDCN-EES-U & 0.676 (+2.58$\%$) & 0.681 (+1.34$\%$) & \textbf{0.536 (+14.78$\%$)} \\
\hline
Dexi & 0.666 & 0.672 & 0.478 \\
\hline
Dexi-ES & 0.671 (+0.75$\%$) & 0.678 (+0.89$\%$) & 0.485 (+1.46$\%$) \\
\hline
Dexi-EES & \textbf{0.683 (+2.55$\%$)} & \textbf{0.691 (+2.83$\%$)} & 0.490 (+2.51$\%$) \\
\hline
Dexi-ES-U & 0.666 (+0.00$\%$) & 0.673 (+0.15$\%$) & \textbf{0.529 (+10.04$\%$)} \\
\hline
Dexi-EES-U & 0.662 (-0.60$\%$) & 0.669 (-0.45$\%$) & 0.521 (+9.00$\%$) \\
\hline
A & 0.657 & 0.667 & 0.454 \\
\hline
A-ES & 0.669 (+1.83$\%$) & 0.675 (+1.20$\%$) & 0.479 (+5.51$\%$) \\
\hline
A-EES & \textbf{0.680 (+3.50$\%$)} & \textbf{0.688 (+3.15$\%$)} & 0.501 (+10.35$\%$) \\
\hline
A-ES-U & 0.669 (+1.83$\%$) & 0.676 (+1.35$\%$) & \textbf{0.533 (+17.40$\%$)} \\
\hline
A-EES-U & 0.669 (+1.83$\%$) & 0.675 (+1.20$\%$) & \textbf{0.533 (+17.40$\%$)} \\
\hline
\end{tabular}
\end{table}

\textbf{Qualitative Analysis}

Figure \ref{Compare} presents visual comparisons of the predicted results. It showcases full images alongside cropped regions (the blue and yellow boxes). The columns represent the original images, ground-truths, and the predictions, respectively. Notably, the proposed EES architecture generates more perceptual edge predictions compared to the baseline model.

\subsection{Ablation studies}

We conduct ablation studies to investigate the impact of different selector structures, positional encoding methods in transformer encoder blocks, and the effect of using multiple extractors and selectors. These experiments further validate that the E-S framework itself is the primary contributor to performance improvements.

\subsubsection{Effect of Positional Encoding}

Table \ref{Pos-Encoding} presents the performance of the standard E-S architecture with various positional encoding strategies applied to the selector. Two types are examined:
\begin{itemize}
    \item Standard positional encoding as in \cite{VS2017Attention}.
    \item Rotary positional encoding as in \cite{SA2024RoFormer}
\end{itemize}

We test both applying positional encoding to all encoder blocks and only the first encoder block of each level. However, the results indicate that positional encoding provides no observed improvement, suggesting that positional information has already been well captured through the CNN backbone of the selector structure. Similar experiments are also implemented on the EES framework, which exhibit consistent results, please refer to the supplementary materials.

\begin{table}[htbp]
\renewcommand\arraystretch{1.5}
\centering
\caption{\qquad \textbf{Standard E-S architecture with positional encoding in the selector:} Results are on BRIND with 1-pixel error tolerance without NMS. Here, \textit{Std} represents applying the standard positional encoding to all encoder blocks, \textit{Rot} represents applying the rotary positional encoding to all encoder blocks, and \textit{Once} represents applying the corresponding positional encoding only in the first encoder block per level rather than in all encoder blocks.}
\label{Pos-Encoding}
\begin{tabular}{|p{28mm}<{\centering}|p{47.75mm}<{\centering}|}
\hline
 &  Applying positional encoding in selectors
\end{tabular}
\begin{tabular}{|p{28mm}<{\centering}|p{13mm}<{\centering}|p{13mm}<{\centering}|p{13mm}<{\centering}|}
\hline
\ Benchmarks   & ODS   & OIS   & AP \\
\hline
HED-ES & 0.668 & 0.677 & 0.466 \\
\hline
Rot-HED-ES & \textbf{0.671} & \textbf{0.680} & 0.474 \\
\hline
Once-Rot-HED-ES & 0.669 & 0.678 & 0.474 \\
\hline
Std-HED-ES & 0.668 & 0.677 & 0.463 \\
\hline
Once-Std-HED-ES & 0.668 & 0.677 & 0.479 \\
\hline
HED-ES-U & 0.668 & 0.673 & 0.522 \\
\hline
Rot-HED-ES-U & 0.665 & 0.672 & 0.532 \\
\hline
Once-Rot-HED-ES-U & 0.663 & 0.670 & \textbf{0.533} \\
\hline
Std-HED-ES-U & 0.667 & 0.674 & 0.526 \\
\hline
Once-Std-HED-ES-U & 0.662 & 0.669 & \textbf{0.533} \\
\hline
BDCN-ES & 0.668 & 0.677 & 0.486 \\
\hline
Rot-BDCN-ES & 0.668 & 0.676 & 0.481 \\
\hline
Once-Rot-BDCN-ES & 0.666 & 0.676 & 0.486 \\
\hline
Std-BDCN-ES & 0.668 & 0.676 & 0.479 \\
\hline
Once-Std-BDCN-ES & 0.667 & 0.676 & 0.484 \\
\hline
BDCN-ES-U & 0.672 & 0.681 & 0.516 \\
\hline
Rot-BDCN-ES-U & 0.652 & 0.660 & 0.510 \\
\hline
Once-Rot-BDCN-ES-U & 0.672 & 0.679 & \textbf{0.515} \\
\hline
Std-BDCN-ES-U & 0.656 & 0.664 & 0.511 \\
\hline
Once-Std-BDCN-ES-U & \textbf{0.679} & \textbf{0.688} & 0.504 \\
\hline
Dexi-ES & \textbf{0.671} & \textbf{0.678} & 0.485 \\
\hline
Rot-Dexi-ES & \textbf{0.671} & 0.677 & 0.492 \\
\hline
Once-Rot-Dexi-ES & 0.670 & 0.677 & 0.485 \\
\hline
Std-Dexi-ES & 0.671 & 0.677 & 0.490 \\
\hline
Once-Std-Dexi-ES & 0.670 & 0.677 & 0.482 \\
\hline
Dexi-ES-U & 0.666 & 0.673 & \textbf{0.529} \\
\hline
Rot-Dexi-ES-U & 0.668 & 0.674 & 0.510 \\
\hline
Once-Rot-Dexi-ES-U & 0.669 & 0.674 & 0.523 \\
\hline
Std-Dexi-ES-U & 0.669 & 0.675 & 0.519 \\
\hline
Once-Std-Dexi-ES-U & 0.667 & 0.673 & 0.512 \\
\hline
\end{tabular}
\end{table}

\subsubsection{Selector Architecture Variants}

Table \ref{Selector-Structure} evaluates the effect of different selector designs within the standard E-S architecture. 

\textbf{We test several variants:}
\begin{itemize}
\item T0: No transformer encoder blocks are employed.

\item T1: Encoder blocks only applied at the $\frac{1}{16}$ scale.

\item T2: Encoder blocks applied at both $\frac{1}{8}$ and $\frac{1}{16}$ scales, namely the selector adopted in the main experimental setup.

\item T2-Double: The same block setting as T2, but with doubled intermediate channel numbers of convolutional layers\footnote{Namely, a convolutional layer outputting n channels is modified to output 2n channels, with the number of input channels of the next layer being modified accordingly. Note that this is not applied to the number of output channels in the first and last convolutional layers.}.

\item T3: Encoder blocks are also applied at $\frac{1}{4}$, besides $\frac{1}{8}$, and $\frac{1}{16}$ scales.

\item T2U: Encoder blocks are moved to $\frac{1}{16}$ and $\frac{1}{8}$ in the up-sampling path, instead of the in the down-sampling path\footnote{Namely, encoder blocks are applied at the $\frac{1}{16}$ scale and the $\frac{1}{8}$ scale before up-sampling to the $\frac{1}{4}$ scale rather than before down-sampling to the $\frac{1}{16}$ scale.}.
\end{itemize}

\begin{table}[htbp]
\renewcommand\arraystretch{1.5}
\centering
\caption{\qquad \textbf{The standard E-S architecture with different selector Structures:} The results are on BRIND with 1-pixel error tolerance without NMS. The notations are the same as in the main-text.}
\label{Selector-Structure}
\begin{tabular}{|p{29mm}<{\centering}|p{47.75mm}<{\centering}|}
\hline
 &  Different selector structures
\end{tabular}
\begin{tabular}{|p{29mm}<{\centering}|p{13mm}<{\centering}|p{13mm}<{\centering}|p{13mm}<{\centering}|}
\hline
  Benchmarks  & ODS   & OIS   & AP \\
\hline
HED-T0-ES & 0.668 & 0.677 & 0.465 \\
\hline
HED-T1-ES & 0.668 & 0.677 & 0.469 \\
\hline
HED-T2-ES & 0.668 & 0.677 & 0.466 \\
\hline
HED-T2-Double-ES & \textbf{0.673} & \textbf{0.682} & 0.483 \\
\hline
HED-T2U-ES & 0.669 & 0.679 & 0.475 \\
\hline
HED-T3-ES & 0.667 & 0.677 & 0.467 \\
\hline
HED-T0-ES-U & 0.662 & 0.670 & 0.526 \\
\hline
HED-T1-ES-U & 0.663 & 0.671 & 0.525 \\
\hline
HED-T2-ES-U & 0.668 & 0.673 & 0.522 \\
\hline
HED-T2-Double-ES-U & 0.662 & 0.670 & \textbf{0.536} \\
\hline
HED-T2U-ES-U & 0.666 & 0.672 & 0.529 \\
\hline
HED-T3-ES-U & 0.663 & 0.671 & 0.525 \\
\hline
BDCN-T0-ES & 0.666 & 0.676 & 0.481 \\
\hline
BDCN-T1-ES & 0.667 & 0.676 & 0.481 \\
\hline
BDCN-T2-ES & 0.668 & 0.677 & 0.486 \\
\hline
BDCN-T2-Double-ES & 0.669 & 0.678 & 0.492 \\
\hline
BDCN-T2U-ES & 0.667 & 0.676 & 0.477 \\
\hline
BDCN-T3-ES & 0.666 & 0.676 & 0.491 \\
\hline
BDCN-T0-ES-U & 0.671 & 0.678 & 0.503 \\
\hline
BDCN-T1-ES-U & 0.662 & 0.671 & \textbf{0.520} \\
\hline
BDCN-T2-ES-U & 0.672 & 0.681 & 0.516 \\
\hline
BDCN-T2-Double-ES-U & \textbf{0.674} & \textbf{0.682} & 0.518 \\
\hline
BDCN-T2U-ES-U & 0.667 & 0.676 & 0.516 \\
\hline
BDCN-T3-ES-U & 0.670 & 0.678 & 0.516 \\
\hline
Dexi-T0-ES & 0.670 & 0.677 & 0.485 \\
\hline
Dexi-T1-ES & 0.670 & 0.677 & 0.483 \\
\hline
Dexi-T2-ES & 0.671 & 0.678 & 0.485 \\
\hline
Dexi-T2-Double-ES & \textbf{0.673} & \textbf{0.680} & 0.493 \\
\hline
Dexi-T2U-ES & 0.671 & 0.677 & 0.482 \\
\hline
Dexi-T3-ES & 0.670 & 0.677 & 0.480 \\
\hline
Dexi-T0-ES-U & 0.668 & 0.674 & 0.511 \\
\hline
Dexi-T1-ES-U & 0.669 & 0.676 & 0.512 \\
\hline
Dexi-T2-ES-U & 0.666 & 0.673 & 0.529 \\
\hline
Dexi-T2-Double-ES-U & 0.669 & 0.676 & \textbf{0.533} \\
\hline
Dexi-T2U-ES-U & 0.667 & 0.674 & 0.509 \\
\hline
Dexi-T3-ES-U & 0.670 & 0.676 & 0.528 \\
\hline
\end{tabular}
\end{table}

The results show that selector architecture contributes marginal impact compared to introducing the E-S framework itself. While enlarging the selector (e.g., employing T2-Double instead of T2) yield modest gains, the standard selector works well. This confirms that the framework design, rather than the specific selector structure, is the dominant factor in performance improvement.

\subsubsection{Multiple Extractors and Multi-Level Selectors}

Table \ref{E-S-Com} shows the effect of using multiple extractors and multiple levels of selection. Here, using multiple extractors represents that the extractors HED, BDCN, and Dexi are combined as a single multi-source extractor, while multi-level selection represents that the E-S or EES architecture is treated as a new extractor, with an additional selector applied on top. For example, 'Com-EES-L2' means applying one additional selector on the normalized output features\footnote{Namely, the features after multiplying weights but before summing, and are normalized by dividing by the maximum value, scaling to the [0,1] range.} from the EES framework, while the extractor in such a situation is the one combining the three extractors\footnote{Namely, the concatenation of their outputted features are treated as the final outputted features.}. The two fixed features, the zeros and the ones, are also being added to features of every level when employing high-level selections via the EES framework.

\textbf{Results show that:} 

\begin{itemize}
    \item In the standard E-S framework, using multiple extractors improves performance. This confirms that features from the standard extractor, namely the outputs of previous models, have been compressed with substantial details lost, and therefore employing multiple extractor scores benefits the final predictions since providing more features allows fewer details lost.
    \item In the EES framework, combining extractors shows only comparable results on ODS/OIS relative to the best single-extractor configuration. This confirms the effectiveness of employing richer and less-loss features as EES does. Namely, features employed in the EES framework are rich enough to provide sufficient options corresponding to the extractor for the selectors, making additional extractors less impactful, although whether the best scores can be obtained depends on whether the best extractor is employed.
    \item In both cases, employing higher-level selections only provides negative effects. This demonstrates that one-level selection is sufficient and more selection can lead to overfitting or diluted features. 
    \item Double the network size of the selector still provides positive results.
\end{itemize}

Importantly, using multiple extractors does not degrade performance significantly, demonstrating the selector’s ability to prioritize useful features. Thus, when the best selection of the extractor is uncertain, combining several and deferring the choice to the selector can be a practical strategy.

\begin{table}[htbp]
\renewcommand\arraystretch{1.5}
\centering
\caption{\qquad \textbf{The proposed frameworks with multiple extractors and multiple-level selections:} The results are on BRIND with 1-pixel error tolerance without NMS. Here, \textit{Com} represents the result related to the combination of the three extractors, HED, BDCN, and Dexi, while \textit{L1}, \textit{L2}, \textit{L3} represents applying the selector once, twice, and three times, respectively. For example, \textit{Com-EES-L2} represents that the final extractor is a combination of the three extractors, and the selector is applied twice, under the EES framework.}
\label{E-S-Com}
\begin{tabular}{|p{28mm}<{\centering}|p{47.75mm}<{\centering}|}
\hline
 &  Multiple extractors and multi-level selectors
\end{tabular}
\begin{tabular}{|p{28mm}<{\centering}|p{13mm}<{\centering}|p{13mm}<{\centering}|p{13mm}<{\centering}|}
\hline
    & ODS   & OIS   & AP \\
\hline
HED-ES & 0.668 & 0.677 & 0.466 \\
\hline
BDCN-ES & 0.668 & 0.677 & 0.486 \\
\hline
Dexi-ES & 0.671 & 0.678 & 0.485 \\
\hline
Com-ES-L1-3 & 0.678 & 0.685 & 0.513 \\
\hline
Com-ES-L2-3 & 0.670 & 0.681 & 0.403 \\
\hline
HED-EES & 0.679 & 0.687 & 0.492 \\
\hline
BDCN-EES & 0.678 & 0.686 & 0.521 \\
\hline
Dexi-EES & 0.683 & 0.691 & 0.490 \\
\hline
Com-EES-L1 & 0.679 & 0.688 & 0.576 \\
\hline
Com-EES-L2 & 0.676 & 0.688 & 0.560 \\
\hline
Com-EES-L3 & 0.675 & 0.687 & 0.536 \\
\hline
Com-EES-Double-L1 & 0.683 & 0.690 & 0.592 \\
\hline
Com-EES-Double-L2 & 0.681 & 0.689 & 0.548 \\
\hline
\end{tabular}
\end{table}

\section{Discussion}

While this work is centered on edge detection, the E-S paradigm is broadly applicable across a wide range of vision tasks. Indeed, results on BSDS500 and NYUD2 datasets demonstrate that the framework could also be suitable for tasks like contour detection and segmentation, please refer to the supplementary materials. Many image-processing challenges benefit from pixel-level feature discrimination, due to the inherently heterogeneous nature of visual data—where structures like edges and textures, high-frequency and low-frequency patterns must be treated differently. Although the research community has extensively explored feature extractors, feature selection remains relatively underdeveloped. We hope this work provides a foundation for advancing this aspect of vision models, encouraging further exploration into robust, generalizable selector designs for a wider range of tasks.

Despite the promising results, several directions remain open for future research:
\begin{itemize}
    \item Specifically-designed Extractors: While we have demonstrated improvements using existing extractors, designing dedicated extractors tailored for E-S or EES frameworks may yield even better performance.
    \item Efficiency Improvements: Enhancing the computational efficiency of the standard and enhanced architectures would increase their viability for real-time or resource-constrained applications, such as high-resolution edge detection in autonomous systems.
    \item Balancing Quantitative and Perceptual Quality: The best quantitative may be achieved without the union training stage, while the most perceptually appealing predictions may be obtained with it. Future work may focus on developing strategies that reconcile this trade-off, achieving both the best metrics and visual quality simultaneously.
\end{itemize}

\section{Conclusion}

In this paper, we introduced a novel Extractor-Selector (E-S) paradigm to enhance image-processing tasks, with a particular focus on edge detection (ED). Unlike prior models that primarily rely on image-level feature fusion, our method emphasizes pixel-wise feature selection, allowing significantly improved performance. 

To address the limitations of the standard E-S framework—specifically its insufficient exploitation of less-loss features—we further proposed the Enhanced Extractor-Selector (EES) framework. By leveraging richer, less-degraded intermediate feature representations and incorporating auxiliary features, the EES architecture delivers more quantitatively accurate and perpetually high-quality edge predictions.

Comprehensive experiments across multiple benchmark datasets demonstrate the substantial performance gains achieved by both the E-S and EES frameworks. Ablation studies further confirm that these improvements are primarily attributed to the framework itself, rather than the specific design of the selector module, highlighting the general utility and robustness of the proposed paradigm. Additionally, the high compatibility of our frameworks with existing models facilitates straightforward adoption and practical deployment, making them a versatile and powerful enhancement strategy for ED systems, while the experimental results on contour detection and segmentation datasets indicate their potential to extend to broader image tasks.

In summary, the proposed E-S and EES frameworks offer a simple yet powerful paradigm that advances the state of edge detection and opens up promising pathways for broader applications in image processing and computer vision.

\section{Bibliography}
\bibliographystyle{unsrt}
\bibliography{EDBitex}

\section{Supplementary Materials}

Supplementary experiment results are presented here, as well as the full details of the selector structure.

\subsection{Results on BSDS500 and NYUD2}

Table \ref{Raw-BSDS} and \ref{Raw-NYU2C} show the experiment results on BSDS500 and NYUD2 datasets.

\begin{table}[htbp]
\renewcommand\arraystretch{1.5}
\centering
\caption{\qquad \textbf{Results on BSDS500 Label 1, with 1-pixel error tolerance, without NMS:} The table is structured as in Table \ref{Raw-BIPED2}, with results for the BSDS500 dataset. In the experiment, only Label 1 of each image is employed as the ground-truth. The dataset is split into 400 training images and 100 testing images, with all training and assessing schemes following the main-text.}
\label{Raw-BSDS}
\begin{tabular}{|p{17mm}<{\centering}|p{62.75mm}<{\centering}|}
\hline
 &  BSDS500 Label 1 without NMS
\end{tabular}
\begin{tabular}{|p{17mm}<{\centering}|p{18mm}<{\centering}|p{18mm}<{\centering}|p{18mm}<{\centering}|}
\hline
    & ODS   & OIS   & AP \\
\hline
HED & 0.420 & 0.433 & 0.175 \\
\hline
HED-ES & 0.466 & 0.479 & 0.253 \\
\hline
HED-EES & \textbf{0.492} & \textbf{0.504} & 0.261 \\
\hline
HED-ES-U & 0.459 & 0.471 & 0.272 \\
\hline
HED-EES-U & 0.484 & 0.493 & \textbf{0.305} \\
\hline
BDCN & 0.454 & 0.465 & 0.240 \\
\hline
BDCN-ES & 0.457 & 0.469 & 0.253  \\
\hline
BDCN-EES & \textbf{0.480} & \textbf{0.494} & 0.266  \\
\hline
BDCN-ES-U & 0.456 & 0.464 & 0.258 \\
\hline
BDCN-EES-U & 0.471 & 0.483 & \textbf{0.288} \\
\hline
Dexi & 0.465 & 0.473 & 0.242 \\
\hline
Dexi-ES & 0.467 & 0.475 & 0.249 \\
\hline
Dexi-EES & \textbf{0.487} & \textbf{0.501} & 0.261 \\
\hline
Dexi-ES-U & 0.473 & 0.484 & 0.271 \\
\hline
Dexi-EES-U & 0.472 & 0.485 & \textbf{0.277} \\
\hline
\end{tabular}
\end{table}

\begin{table}[htbp]
\renewcommand\arraystretch{1.5}
\centering
\caption{\qquad \textbf{Results on NYU2C, with 1-pixel error tolerance, without NMS:} The table is structured as in Table \ref{Raw-BIPED2}, with results for the NYUD2 dataset. The edge label (ground-truth) of each image is extracted by using the Canny operator from the segmentation label. The dataset is split into 1100 training images and 300 testing images, from the 1449 annotated data, with all training and assessing schemes following the main-text.}
\label{Raw-NYU2C}
\begin{tabular}{|p{18mm}<{\centering}|p{59.75mm}<{\centering}|}
\hline
 &  NYU2C without NMS
\end{tabular}
\begin{tabular}{|p{18mm}<{\centering}|p{17mm}<{\centering}|p{17mm}<{\centering}|p{17mm}<{\centering}|}
\hline
    & ODS   & OIS   & AP \\
\hline
HED & 0.376 & 0.386 & 0.146 \\
\hline
HED-ES & 0.396 & 0.406 & 0.158 \\
\hline
HED-EES & 0.409 & \textbf{0.418} & 0.154 \\
\hline
HED-ES-U & 0.402 & 0.411 & \textbf{0.172} \\
\hline
HED-EES-U & \textbf{0.410} & \textbf{0.418} & 0.171 \\
\hline
BDCN & 0.406 & 0.414 & \textbf{0.183} \\
\hline
BDCN-ES & 0.403 & 0.411 & 0.175  \\
\hline
BDCN-EES & \textbf{0.408} & \textbf{0.417} & 0.142  \\
\hline
BDCN-ES-U & 0.398 & 0.408 & 0.166 \\
\hline
BDCN-EES-U & 0.403 & 0.412 & 0.169 \\
\hline
Dexi & 0.408 & 0.416 & 0.161 \\
\hline
Dexi-ES & 0.406 & 0.414 & 0.160 \\
\hline
Dexi-EES & 0.402 & 0.413 & 0.158 \\
\hline
Dexi-ES-U & 0.414 & 0.422 & \textbf{0.181} \\
\hline
Dexi-EES-U & \textbf{0.417} & \textbf{0.427} & 0.174 \\
\hline
\end{tabular}
\end{table}

\subsection{Evaluation Results Under traditional relaxed settings}

 Table \ref{BIPED},\ref{UDED},\ref{BRIND},\ref{BSDS},\ref{NYU2C} display the evaluation result under traditional relaxed settings, namely 0.0075 error tolerance set in the algorithm presented in \cite{MF2004Learning} and with NMS. The setting results in error tolerances about 11.1 pixels for BIPED2, 4.3 pixels for BRIND and BSDS500, and 6 pixels for NYUD2, respectively.
 
\begin{table}[htbp]
\renewcommand\arraystretch{1.5}
\centering
\caption{\qquad \textbf{Results on BIPED2 with 11.1-pixel error tolerance, with NMS:} The table is structured as in Table \ref{Raw-BIPED2}. These are the results under the traditional evaluation settings that the error tolerance distance in the evaluation algorithm is set to 0.0075.}
\label{BIPED}
\begin{tabular}{|p{18mm}<{\centering}|p{59.75mm}<{\centering}|}
\hline
 &  BIPED2 with NMS
\end{tabular}
\begin{tabular}{|p{18mm}<{\centering}|p{17mm}<{\centering}|p{17mm}<{\centering}|p{17mm}<{\centering}|}
\hline
    & ODS   & OIS   & AP \\
\hline
HED & 0.883 & 0.894 & 0.914 \\
\hline
HED-ES & 0.878 & 0.889 & 0.926 \\
\hline
HED-EES & 0.882 & 0.893 & \textbf{0.927} \\
\hline
HED-ES-U & 0.878 & 0.890 & 0.908 \\
\hline
HED-EES-U & \textbf{0.884} & \textbf{0.892} & 0.900 \\
\hline
BDCN & 0.880 & 0.891 & 0.912 \\
\hline
BDCN-ES & 0.879 & 0.890 & 0.918 \\
\hline
BDCN-EES & 0.881 & 0.890 & 0.921 \\
\hline
BDCN-ES-U & 0.876 & 0.886 & \textbf{0.922} \\
\hline
BDCN-EES-U & \textbf{0.884} & \textbf{0.891} & 0.860 \\
\hline
Dexi & \textbf{0.890} & \textbf{0.897} & 0.923 \\
\hline
Dexi-ES & 0.889 & \textbf{0.897} & 0.928 \\
\hline
Dexi-EES & 0.887 & 0.895 & \textbf{0.934} \\
\hline
Dexi-ES-U & 0.887 & 0.895 & 0.898 \\
\hline
Dexi-EES-U & 0.881 & 0.890 & 0.887 \\
\hline
\end{tabular}
\end{table}

\begin{table}[htbp]
\renewcommand\arraystretch{1.5}
\centering
\caption{\qquad \textbf{Results on UDED with 0.0075 error tolerance, with NMS.} The table is structured as in Table \ref{Raw-BIPED2}. These are the results under the traditional evaluation settings that the error tolerance distance in the evaluation algorithm is set to 0.0075.}
\label{UDED}
\begin{tabular}{|p{18mm}<{\centering}|p{59.75mm}<{\centering}|}
\hline
 &  UDED with NMS
\end{tabular}
\begin{tabular}{|p{18mm}<{\centering}|p{17mm}<{\centering}|p{17mm}<{\centering}|p{17mm}<{\centering}|}
\hline
    & ODS   & OIS   & AP \\
\hline
HED & 0.819 & 0.857 & 0.806 \\
\hline
HED-ES & 0.820 & 0.852 & 0.823 \\
\hline
HED-EES & 0.823 & 0.845 & \textbf{0.856 }\\
\hline
HED-ES-U & 0.794 & 0.829 & 0.799  \\
\hline
HED-EES-U & \textbf{0.833} & \textbf{0.861} & 0.802  \\
\hline
BDCN& 0.798 & 0.837 & 0.796 \\
\hline
BDCN-ES& 0.809 & 0.843 & 0.812 \\
\hline
BDCN-EES& 0.825 & 0.854 & 0.822 \\
\hline
BDCN-ES-U& 0.834 & 0.858 & 0.790 \\
\hline
BDCN-EES-U& \textbf{0.841} & \textbf{0.859} & \textbf{0.852} \\
\hline
Dexi& 0.829 & 0.858 & 0.824 \\
\hline
Dexi-ES& 0.829 & 0.859 & 0.829 \\
\hline
Dexi-EES& 0.852 & 0.869 & \textbf{0.889} \\
\hline
Dexi-ES-U& 0.847 & 0.865 & 0.793 \\
\hline
Dexi-EES-U& \textbf{0.853} & \textbf{0.874} & 0.767 \\
\hline
\end{tabular}
\end{table}

\begin{table}[htbp]
\renewcommand\arraystretch{1.5}
\centering
\caption{\qquad \textbf{Results on BRIND with 4.3-pixel error tolerance, with NMS:} The table is structured as in Table \ref{Raw-BIPED2}. These are the results under the traditional evaluation settings that the error tolerance distance in the evaluation algorithm is set to 0.0075.}
\label{BRIND}
\begin{tabular}{|p{18mm}<{\centering}|p{59.75mm}<{\centering}|}
\hline
 &  BRIND with NMS
\end{tabular}
\begin{tabular}{|p{18mm}<{\centering}|p{17mm}<{\centering}|p{17mm}<{\centering}|p{17mm}<{\centering}|}
\hline
    & ODS   & OIS   & AP \\
\hline
HED & 0.784 & 0.799 & 0.832 \\
\hline
HED-ES & 0.787 & 0.800 & 0.843 \\
\hline
HED-EES & \textbf{0.789} & \textbf{0.803} & \textbf{0.849} \\
\hline
HED-ES-U & 0.785 & 0.796 & 0.799 \\
\hline
HED-EES-U & 0.788 & 0.801 & 0.806 \\
\hline
BDCN & 0.775 & 0.791 & 0.806 \\
\hline
BDCN-ES & 0.781 & 0.795 & \textbf{0.832}  \\
\hline
BDCN-EES & \textbf{0.790} & \textbf{0.806} & 0.831  \\
\hline
BDCN-ES-U & 0.786 & 0.799 & 0.828 \\
\hline
BDCN-EES-U & 0.788 & 0.798 & 0.820 \\
\hline
Dexi & \textbf{0.794} & \textbf{0.809} & 0.828 \\
\hline
Dexi-ES & 0.792 & 0.806 & 0.837 \\
\hline
Dexi-EES & 0.792 & 0.806 & \textbf{0.842} \\
\hline
Dexi-ES-U & 0.789 & 0.803 & 0.795 \\
\hline
Dexi-EES-U & 0.783 & 0.796 & 0.794 \\
\hline
\end{tabular}
\end{table}

\begin{table}[htbp]
\renewcommand\arraystretch{1.5}
\centering
\caption{\qquad \textbf{Results on BSDS500 Label 1, with 4.3-pixel error tolerance, with NMS:} The table is structured as in Table \ref{Raw-BIPED2}. These are the results under the traditional evaluation settings that the error tolerance distance in the evaluation algorithm is set to 0.0075.}
\label{BSDS}
\begin{tabular}{|p{18mm}<{\centering}|p{59.75mm}<{\centering}|}
\hline
 &  BSDS500 Label 1 with NMS
\end{tabular}
\begin{tabular}{|p{18mm}<{\centering}|p{17mm}<{\centering}|p{17mm}<{\centering}|p{17mm}<{\centering}|}
\hline
    & ODS   & OIS   & AP \\
\hline
HED & 0.622 & 0.650 & 0.578 \\
\hline
HED-ES & 0.628 & 0.654 & 0.611 \\
\hline
HED-EES & \textbf{0.651} & \textbf{0.674} & \textbf{0.661} \\
\hline
HED-ES-U & 0.625 & 0.649 & 0.564 \\
\hline
HED-EES-U & 0.647 & 0.671 & 0.618 \\
\hline
BDCN & 0.628 & 0.650 & 0.596 \\
\hline
BDCN-ES & 0.628 & 0.649 & 0.602  \\
\hline
BDCN-EES & \textbf{0.640} & \textbf{0.666} & \textbf{0.641}  \\
\hline
BDCN-ES-U & 0.617 & 0.640 & 0.570 \\
\hline
BDCN-EES-U & 0.636 & 0.659 & 0.601 \\
\hline
Dexi & 0.648 & 0.672 & 0.608 \\
\hline
Dexi-ES & 0.647 & 0.671 & 0.626 \\
\hline
Dexi-EES & \textbf{0.649} & 0.676 & \textbf{0.654} \\
\hline
Dexi-ES-U & 0.646 & \textbf{0.678} & 0.596 \\
\hline
Dexi-EES-U & 0.642 & 0.674 & 0.608 \\
\hline
\end{tabular}
\end{table}

\begin{table}[htbp]
\renewcommand\arraystretch{1.5}
\centering
\caption{\qquad \textbf{Results on NYU2C, with 6-pixel error tolerance, with NMS:} The table is structured as in Table \ref{Raw-BIPED2}. These are the results under the traditional evaluation settings that the error tolerance distance in the evaluation algorithm is set to 0.0075.}
\label{NYU2C}
\begin{tabular}{|p{18mm}<{\centering}|p{59.75mm}<{\centering}|}
\hline
 &  NYU2C with NMS
\end{tabular}
\begin{tabular}{|p{18mm}<{\centering}|p{17mm}<{\centering}|p{17mm}<{\centering}|p{17mm}<{\centering}|}
\hline
    & ODS   & OIS   & AP \\
\hline
HED & 0.703 & 0.716 & 0.694 \\
\hline
HED-ES & 0.712 & 0.724 & \textbf{0.712} \\
\hline
HED-EES & 0.710 & 0.721 & 0.687 \\
\hline
HED-ES-U & 0.715 & 0.725 & 0.674 \\
\hline
HED-EES-U & \textbf{0.723} & \textbf{0.733} & 0.682 \\
\hline
BDCN & 0.711 & 0.723 & 0.712 \\
\hline
BDCN-ES & 0.711 & 0.723 & \textbf{0.713}  \\
\hline
BDCN-EES & \textbf{0.720} & \textbf{0.732} & 0.708  \\
\hline
BDCN-ES-U & 0.715 & 0.725 & 0.678 \\
\hline
BDCN-EES-U & 0.718 & 0.728 & 0.684 \\
\hline
Dexi & 0.715 & 0.725 & \textbf{0.728} \\
\hline
Dexi-ES & 0.714 & 0.724 & 0.723 \\
\hline
Dexi-EES & 0.634 & 0.644 & 0.652 \\
\hline
Dexi-ES-U & 0.725 & 0.735 & 0.685 \\
\hline
Dexi-EES-U & \textbf{0.728} & \textbf{0.738} & 0.688 \\
\hline
\end{tabular}
\end{table}

\subsection{Results on RCF}

Since RCF can be viewed as an extended version of HED by employing richer features from different layers of the same scales, using it as the extractor on the EES framework is not valid since the EES framework of HED has already employed richer features. Here, we only display the experimental results with the standard E-S framework, please see Table \ref{Raw-RCF} and \ref{RCF}. The E-S framework obtains the best results for most cases, especially under the suggested strict 1-pixel error tolerance.

\begin{table}[htbp]
\renewcommand\arraystretch{1.5}
\centering
\caption{\qquad \textbf{Results on RCF, with 1-pixel error tolerance, without NMS:} The table is structured as in Table \ref{Raw-BIPED2}.}
\label{Raw-RCF}
\begin{tabular}{|p{18mm}<{\centering}|p{59.75mm}<{\centering}|}
\hline
 &  RCF on BIPED2 without NMS
\end{tabular}
\begin{tabular}{|p{18mm}<{\centering}|p{17mm}<{\centering}|p{17mm}<{\centering}|p{17mm}<{\centering}|}
\hline
    & ODS   & OIS   & AP \\
\hline
RCF   & 0.591 & 0.597 & 0.345  \\
\hline
RCF-S & 0.641 & 0.645 & 0.418 \\
\hline
RCF-S-U& \textbf{0.666} & \textbf{0.672} & \textbf{0.502} \\
\hline
\end{tabular}
\begin{tabular}{|p{18mm}<{\centering}|p{59.75mm}<{\centering}|}
\hline
 &  RCF on UDED without NMS
\end{tabular}
\begin{tabular}{|p{18mm}<{\centering}|p{17mm}<{\centering}|p{17mm}<{\centering}|p{17mm}<{\centering}|}
\hline
    & ODS   & OIS   & AP \\
\hline
RCF & 0.699 & 0.741 & 0.450 \\
\hline
RCF-S & 0.723 & 0.761 & 0.562 \\
\hline
RCF-S-U & \textbf{0.752} & \textbf{0.781} & \textbf{0.635} \\
\hline
\end{tabular}
\begin{tabular}{|p{18mm}<{\centering}|p{59.75mm}<{\centering}|}
\hline
 &  RCF on BRIND without NMS
\end{tabular}
\begin{tabular}{|p{18mm}<{\centering}|p{17mm}<{\centering}|p{17mm}<{\centering}|p{17mm}<{\centering}|}
\hline
    & ODS   & OIS   & AP \\
\hline
RCF & 0.646 & 0.656 & 0.394 \\
\hline
RCF-ES & \textbf{0.667} & \textbf{0.675} & 0.475 \\
\hline
RCF-ES-U & 0.666 & 0.672 & \textbf{0.517} \\
\hline
\end{tabular}
\begin{tabular}{|p{18mm}<{\centering}|p{59.75mm}<{\centering}|}
\hline
 &  RCF on BSDS500 Label 1 without NMS
\end{tabular}
\begin{tabular}{|p{18mm}<{\centering}|p{17mm}<{\centering}|p{17mm}<{\centering}|p{17mm}<{\centering}|}
\hline
    & ODS   & OIS   & AP \\
\hline
RCF & 0.426 & 0.441 & 0.190  \\
\hline
RCF-S & 0.459 & 0.472 & 0.239 \\
\hline
RCF-S-U & \textbf{0.461} & \textbf{0.474} & \textbf{0.274} \\
\hline
\end{tabular}
\begin{tabular}{|p{18mm}<{\centering}|p{59.75mm}<{\centering}|}
\hline
 &  RCF on NYUD2 without NMS
\end{tabular}
\begin{tabular}{|p{18mm}<{\centering}|p{17mm}<{\centering}|p{17mm}<{\centering}|p{17mm}<{\centering}|}
\hline
    & ODS   & OIS   & AP \\
\hline
RCF & 0.374 & 0.382 & 0.149  \\
\hline
RCF-S & 0.392 & 0.400 & 0.154 \\
\hline
RCF-S-U & \textbf{0.398} & \textbf{0.407} & \textbf{0.157} \\
\hline
\end{tabular}
\end{table}

\begin{table}[htbp]
\renewcommand\arraystretch{1.5}
\centering
\caption{\qquad \textbf{Results on RCF, with 0.0075 error tolerance, with NMS:} The table is structured as in Table \ref{Raw-BIPED2}.}
\label{RCF}
\begin{tabular}{|p{18mm}<{\centering}|p{59.75mm}<{\centering}|}
\hline
 &  RCF on BIPED2 with NMS
\end{tabular}
\begin{tabular}{|p{18mm}<{\centering}|p{17mm}<{\centering}|p{17mm}<{\centering}|p{17mm}<{\centering}|}
\hline
    & ODS   & OIS   & AP \\
\hline
RCF & \textbf{0.887} & \textbf{0.895} & 0.914  \\
\hline
RCF-ES & 0.882 & 0.891 & \textbf{0.927} \\
\hline
RCF-ES-U & 0.881 & 0.890 & 0.901 \\
\hline
\end{tabular}
\begin{tabular}{|p{18mm}<{\centering}|p{59.75mm}<{\centering}|}
\hline
 &  RCF on UDED with NMS
\end{tabular}
\begin{tabular}{|p{18mm}<{\centering}|p{17mm}<{\centering}|p{17mm}<{\centering}|p{17mm}<{\centering}|}
\hline
    & ODS   & OIS   & AP \\
\hline
RCF& 0.794 & 0.843 & 0.775 \\
\hline
RCF-ES& 0.799 & 0.845 & \textbf{0.833} \\
\hline
RCF-ES-U & \textbf{0.825} & \textbf{0.852} & 0.805 \\
\hline
\end{tabular}
\begin{tabular}{|p{18mm}<{\centering}|p{59.75mm}<{\centering}|}
\hline
 &  RCF on BRIND with NMS
\end{tabular}
\begin{tabular}{|p{18mm}<{\centering}|p{17mm}<{\centering}|p{17mm}<{\centering}|p{17mm}<{\centering}|}
\hline
    & ODS   & OIS   & AP \\
\hline
RCF & \textbf{0.786} & \textbf{0.800} & 0.836  \\
\hline
RCF-ES & 0.785 & 0.797 & \textbf{0.839} \\
\hline
RCF-ES-U & 0.783 & 0.793 & 0.796 \\
\hline
\end{tabular}
\begin{tabular}{|p{18mm}<{\centering}|p{59.75mm}<{\centering}|}
\hline
 &  RCF on BSDS500 Label 1 with NMS
\end{tabular}
\begin{tabular}{|p{18mm}<{\centering}|p{17mm}<{\centering}|p{17mm}<{\centering}|p{17mm}<{\centering}|}
\hline
    & ODS   & OIS   & AP \\
\hline
RCF & 0.624 & \textbf{0.655} & 0.575  \\
\hline
RCF-ES & 0.626 & \textbf{0.655} & \textbf{0.595} \\
\hline
RCF-ES-U & \textbf{0.629} & 0.651 & 0.558 \\
\hline
\end{tabular}
\begin{tabular}{|p{18mm}<{\centering}|p{59.75mm}<{\centering}|}
\hline
 &  RCF on NYUD2 with NMS
\end{tabular}
\begin{tabular}{|p{18mm}<{\centering}|p{17mm}<{\centering}|p{17mm}<{\centering}|p{17mm}<{\centering}|}
\hline
    & ODS   & OIS   & AP \\
\hline
RCF & 0.702 & 0.714 & 0.660  \\
\hline
RCF-ES & 0.712 & 0.724 & \textbf{0.679} \\
\hline
RCF-ES-U & \textbf{0.719} & \textbf{0.728} & 0.673 \\
\hline
\end{tabular}
\end{table}

\subsection{Experiments of Positional Encoding of EES}

Table \ref{EES-Pos-Encoding} exhibits the performance of the EES architecture with various positional encoding strategies applied to the selector. Also, applying positional encoding provides no benefits mostly.

\begin{table}[htbp]
\renewcommand\arraystretch{1.5}
\centering
\caption{\qquad \textbf{EES architecture with positional encoding in the selector:} Results are on BRIND with 1-pixel error tolerance without NMS. The notations are the same as in the main-text.}
\label{EES-Pos-Encoding}
\begin{tabular}{|p{30mm}<{\centering}|p{47.75mm}<{\centering}|}
\hline
 &  Applying positional encoding in selectors
\end{tabular}
\begin{tabular}{|p{30mm}<{\centering}|p{13mm}<{\centering}|p{13mm}<{\centering}|p{13mm}<{\centering}|}
\hline
\ Benchmarks   & ODS   & OIS   & AP \\
\hline
HED-EES & \textbf{0.679} & 0.687 & 0.492 \\
\hline
Rot-HED-EES & \textbf{0.679} & \textbf{0.689} & 0.517 \\
\hline
Once-Rot-HED-EES & 0.636 & 0.646 & 0.435 \\
\hline
Std-HED-EES & 0.678 & 0.685 & 0.522 \\
\hline
Once-Std-HED-EES & 0.674 & 0.684 & 0.508 \\
\hline
HED-EES-U & 0.668 & 0.673 & 0.522 \\
\hline
Rot-HED-EES-U & 0.667 & 0.675 & 0.539 \\
\hline
Once-Rot-HED-EES-U & 0.665 & 0.673 & 0.529 \\
\hline
Std-HED-EES-U & 0.664 & 0.671 & \textbf{0.530} \\
\hline
Once-Std-HED-EES-U & 0.668 & 0.675 & 0.504 \\
\hline
BDCN-EES & \textbf{0.678} & 0.686 & \textbf{0.521} \\
\hline
Rot-BDCN-EES & 0.677 & 0.686 & 0.519 \\
\hline
Once-Rot-BDCN-EES & \textbf{0.678} & \textbf{0.687} & 0.503 \\
\hline
Std-BDCN-EES & \textbf{0.678} & 0.686 & 0.516 \\
\hline
Once-Std-BDCN-EES & \textbf{0.678} & \textbf{0.687} & 0.513 \\
\hline
BDCN-EES-U & 0.672 & 0.681 & 0.516 \\
\hline
Rot-BDCN-EES-U & 0.672 & 0.680 & 0.535 \\
\hline
Once-Rot-BDCN-EES-U & 0.660 & 0.667 & 0.507 \\
\hline
Std-BDCN-EES-U & 0.656 & 0.662 & 0.502 \\
\hline
Once-Std-BDCN-EES-U & 0.667 & 0.675 & 0.486 \\
\hline
Dexi-EES & \textbf{0.683} & \textbf{0.691} & 0.490 \\
\hline
Rot-Dexi-EES & 0.647 & 0.658 & 0.424 \\
\hline
Once-Rot-Dexi-EES & 0.644 & 0.654 & 0.433 \\
\hline
Std-Dexi-EES & 0.682 & 0.690 & 0.479 \\
\hline
Once-Std-Dexi-EES & 0.680 & 0.688 & 0.496 \\
\hline
Dexi-EES-U & 0.666 & 0.673 & \textbf{0.529} \\
\hline
Rot-Dexi-EES-U & 0.666 & 0.672 & 0.517 \\
\hline
Once-Rot-Dexi-EES-U & 0.664 & 0.672 & 0.526 \\
\hline
Std-Dexi-EES-U & 0.664 & 0.672 & 0.522 \\
\hline
Once-Std-Dexi-EES-U & 0.657 & 0.664 & 0.506 \\
\hline
\end{tabular}
\end{table}

\subsection{Detailed Descriptions of the Selector}

\begin{figure*}[htbp]
	\begin{center}
		\subfigure
		[\textbf{An overview architecture of the feature selector}]
		{
			\includegraphics[width=1.8\columnwidth]{Selecting_Model.png}
			\label{Selecting-Model}
		}
		\hspace{0.1in}
		
		\subfigure
		[\textbf{FEB}]
		{
			\includegraphics[width=0.25\columnwidth]{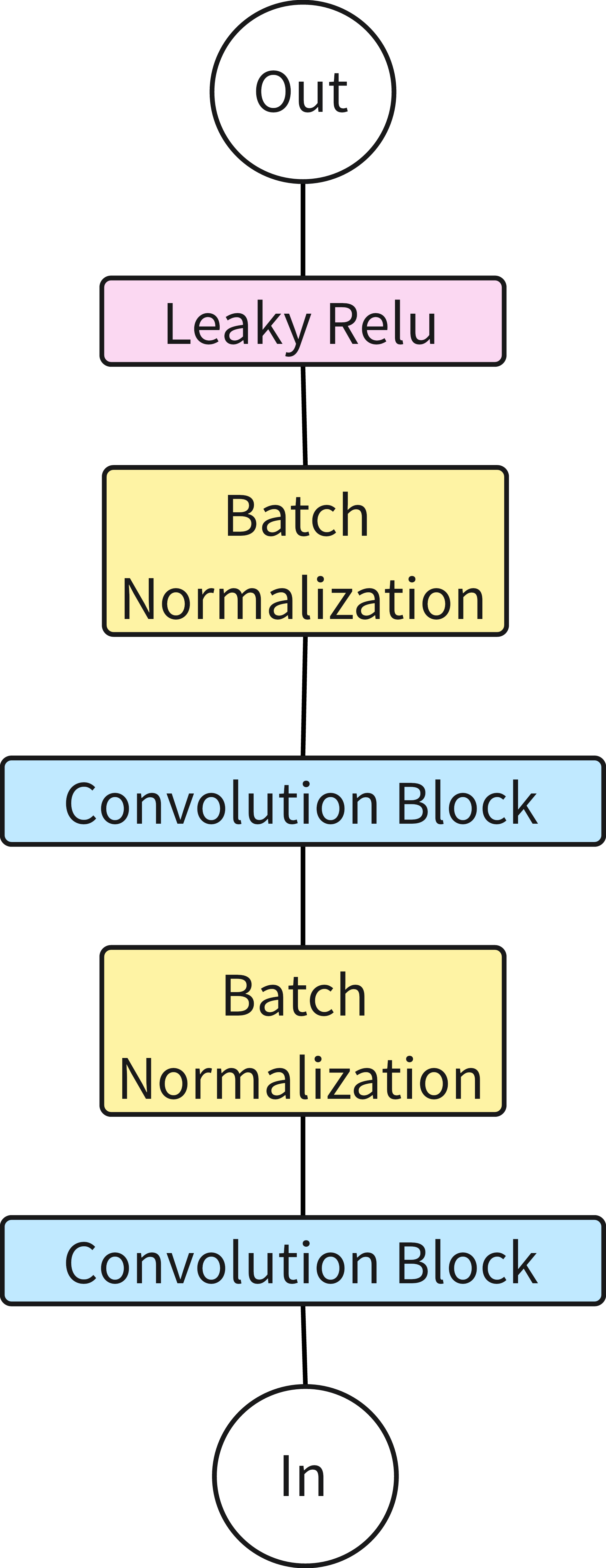}
			\label{Feature-Extract-Block}
		}
		\hspace{0.1in}
		\subfigure
		[\textbf{DSB}]
		{
			\includegraphics[width=0.25\columnwidth]{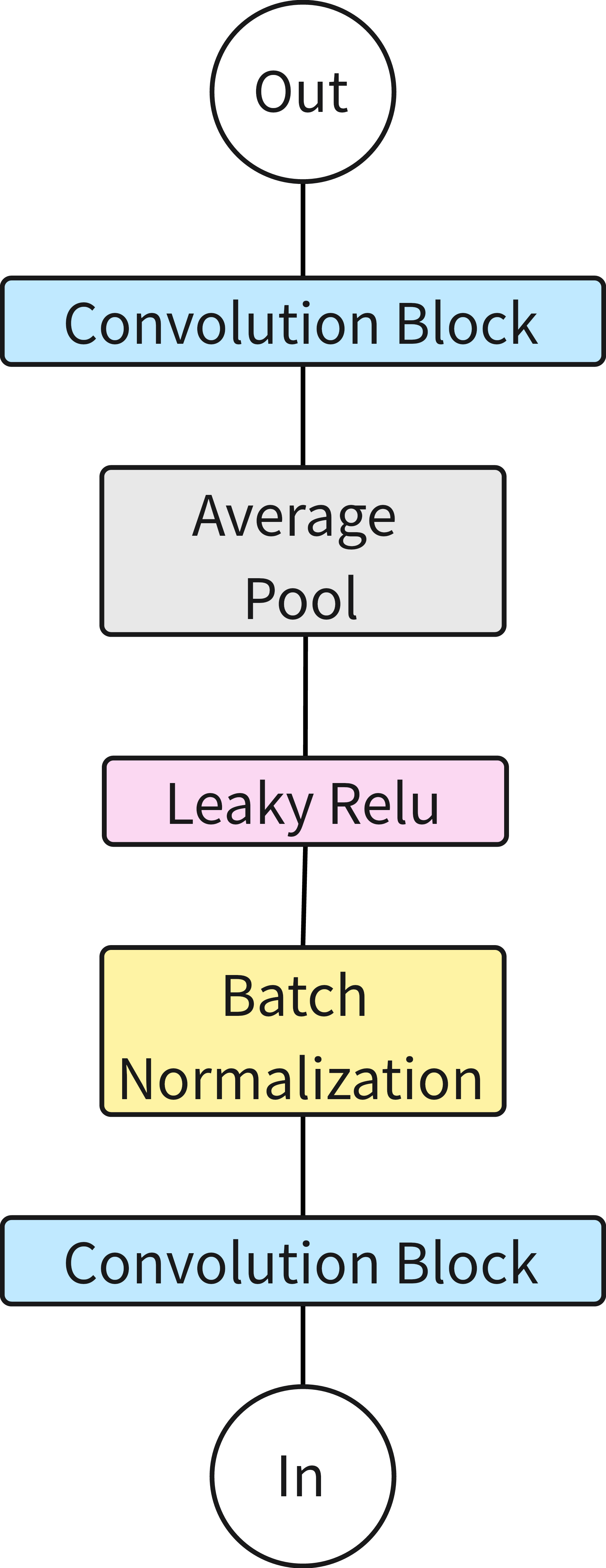}
			\label{Down-Sample-Block}
		}
		\hspace{0.1in}
		\subfigure
		[\textbf{USB}]
		{
			\includegraphics[width=0.25\columnwidth]{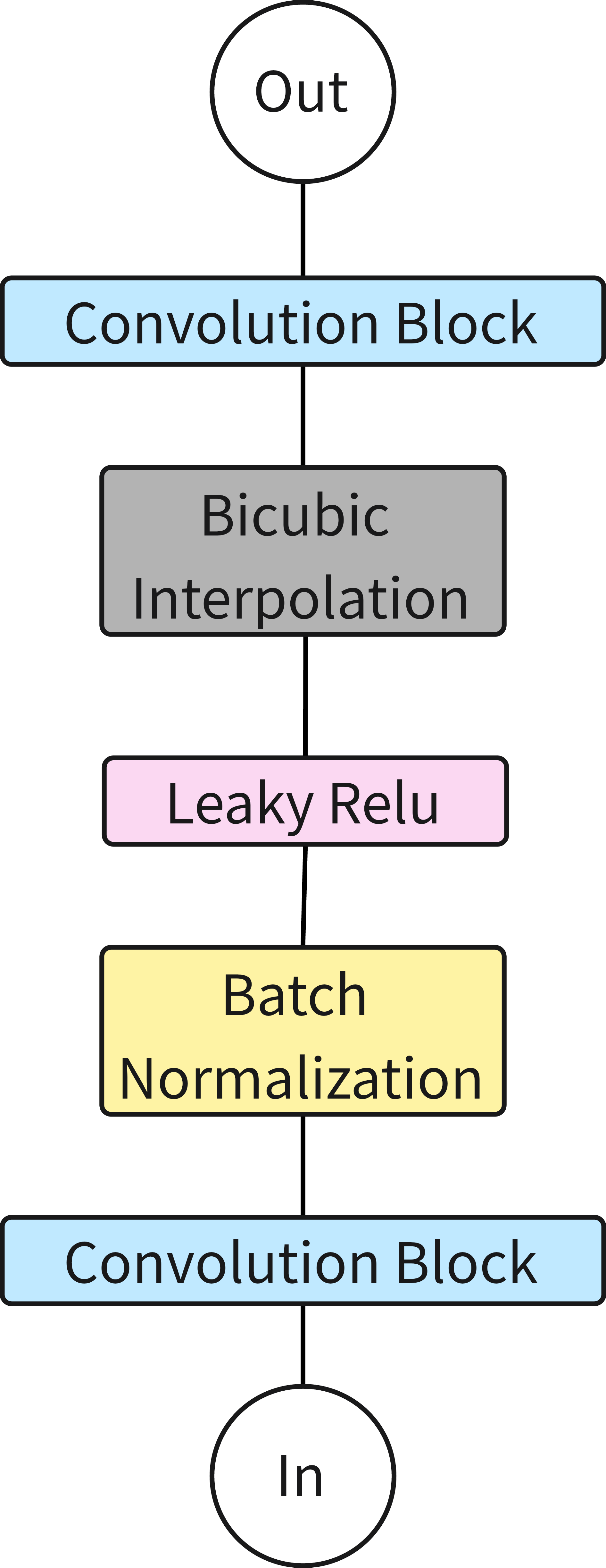}
			\label{Up-Sample-Block}
		}
		\hspace{0.1in}
		\subfigure
		[\textbf{WFB}]
		{
			\includegraphics[width=0.2\columnwidth]{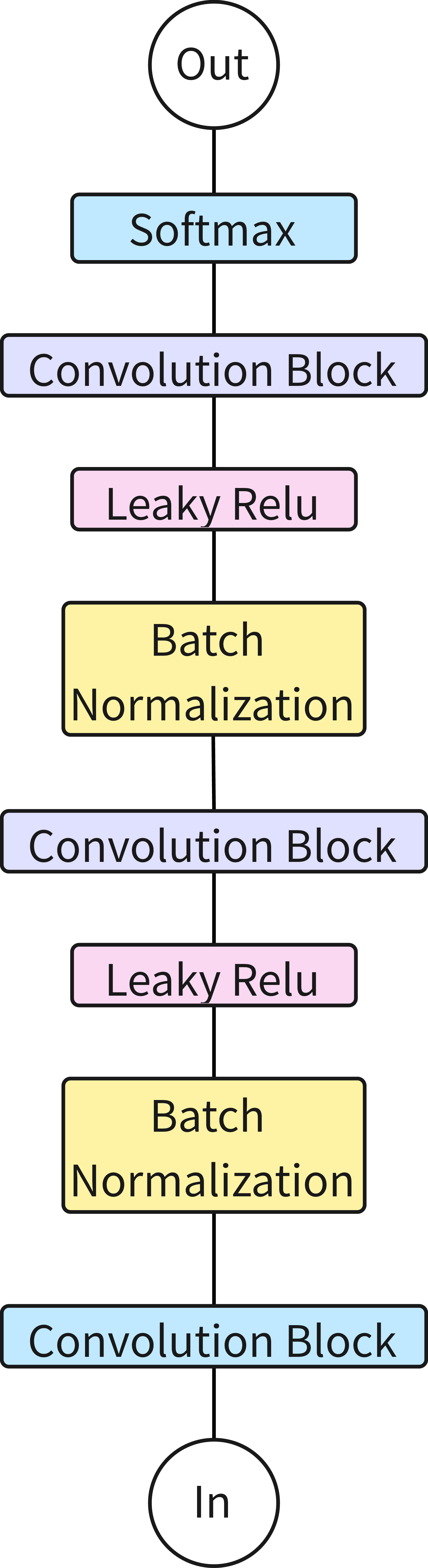}
			\label{Weight-Fuse-Block}
		}
		\hspace{0.1in}
		\subfigure
		[\textbf{EB}]
		{
			\includegraphics[width=0.15\columnwidth]{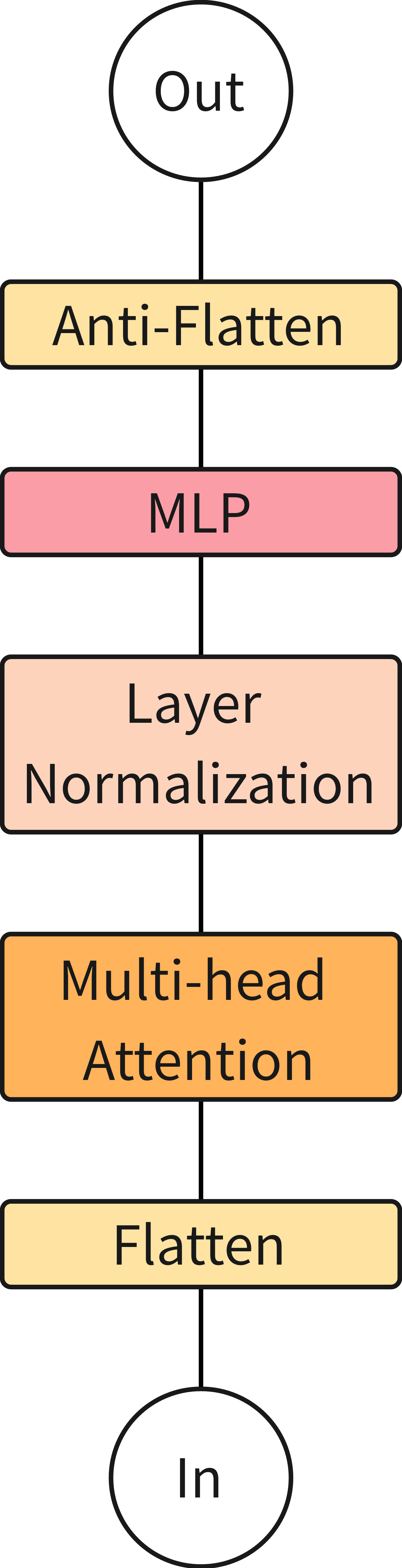}
			\label{Encoder-Block}
		}
		\hspace{0.1in}
		\subfigure
		[\textbf{MLP}]
		{
			\includegraphics[width=0.25\columnwidth]{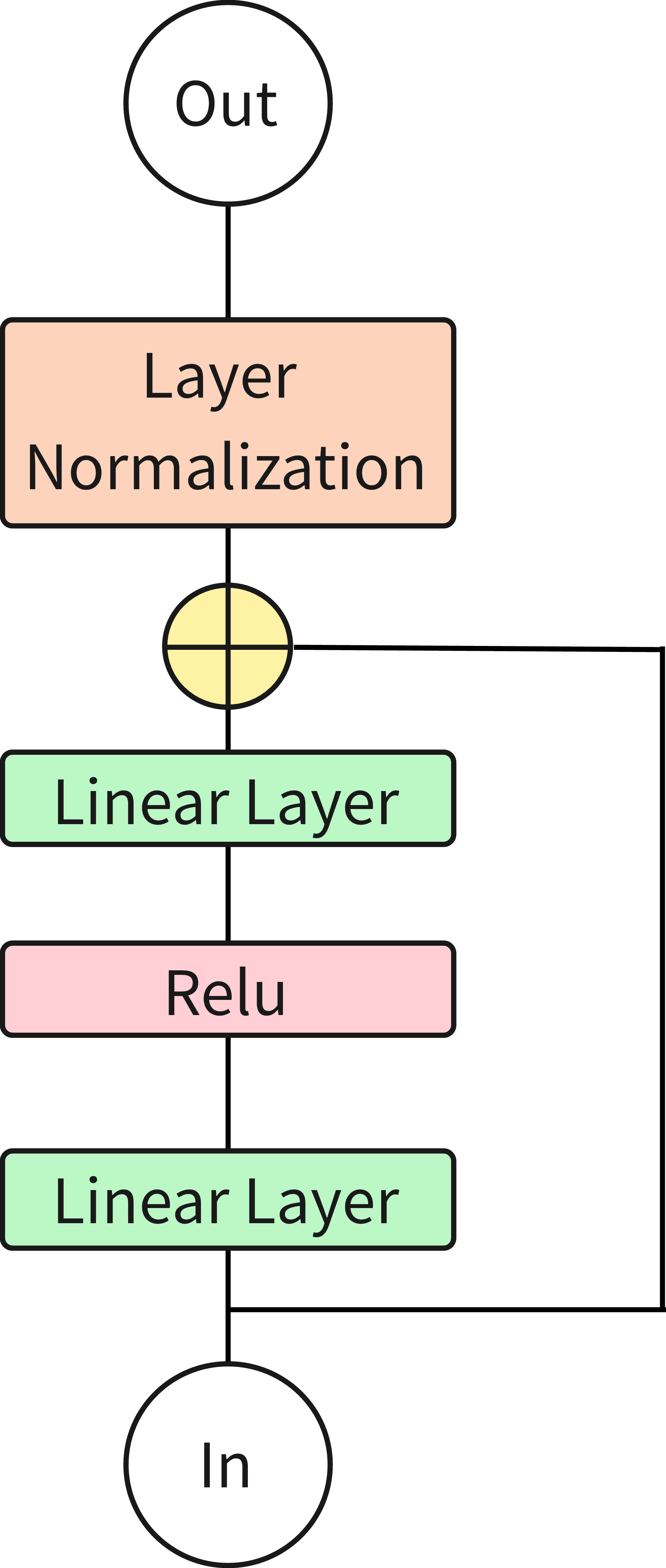}
			\label{MLP}
		}
		
		\caption{\quad	\textbf{The architecture of the feature selector.}
			\\
			\textbf{Figure \ref{Selecting-Model}} provides an overview of the feature selector. It is structured like a U-Net. The input images pass through a feature extraction block, followed by four $\frac{1}{2}$ down-sampling blocks to capture multi-scale features. At the $\frac{1}{8}$ and $\frac{1}{16}$ scales, cascaded transformer encoder blocks are applied. The features from the $\frac{1}{16}$ scale are progressively up-sampled, and each up-sample block incorporates a residual connection with a learnable coefficient to balance the up-sampled features and the features from the residual connections. Finally, pixel-wise weights are produced by fusing the ordinary-scale features.
			\\
			\textbf{Figure \ref{Feature-Extract-Block} (FEB)} is the \textbf{Feature Extract Block}. It consists of two convolution layers, each followed by a batch normalization layer, and an activation function using Leaky ReLU. The convolution layers are $3 \times 3$ in size with each outputs 32 channels typically.
			\\
			\textbf{Figure \ref{Down-Sample-Block} (DSB)} is the \textbf{Down-Sample Block}. The down-sample operation applies a $3 \times 3$ average-pool operation with a stride of 2. The input features go through a $3 \times 3$ convolution layer that doubles the number of channels, followed by batch normalization, Leaky ReLU activation, and average pooling. Another convolution layer maintaining the number of input and output channels is then applied.
			\\
			\textbf{Figure \ref{Up-Sample-Block} (USB)} is the \textbf{Up-Sample Block}. It replaces the average-pool operation from the down-sample block with a $\times 2$ bi-cubic interpolation operation. The remaining structure is similar to the down-sample block. The convolution layers are also $3 \times 3$ in size, but the first one maintains the number of channels of the input and output while the second one halvers the number of channels.
			\\
			\textbf{Figure \ref{Weight-Fuse-Block} (WFB)} is the final \textbf{Weight Fusing Block}. It includes a $3 \times 3$ convolution, batch normalization, and Leaky ReLU activation as the first group. A second structure is similar but with the $1 \times 1$ convolution instead of a $3 \times 3$ one. Finally, another $1 \times 1$ convolution is applied, followed by the Softmax function to produce pixel-wise weights. Typically, the number of output channels of the three convolutional layers is 64, 128, and the required number of output channels of the selector, respectively.
			\\
			\textbf{Figure \ref{Encoder-Block} (EB)} is the \textbf{Encoder Block} from transformers. It includes a flattening operation, a multi-head attention layer, layer normalization, and an MLP. The flattening operation transforms 2D images into 1D vectors for processing in the multi-head attention layer, which typically uses 8 attention heads. The MLP further processes the features, and the resulting features are then reshaped back into 2D images.
			\\
			\textbf{Figure \ref{MLP} (MLP)} is the Multiple Layer Perception (\textbf{MLP}). It consists of two linear layers with ReLU activation, a residual connection, and layer normalization. The numbers of input and output states are the same, while the number of middle states is four times as it.
		}
		\label{Full_Selector}
	\end{center}
\end{figure*}

The details of the blocks employed in the selector are displayed in Figure \ref{Full_Selector} 

\end{document}